\documentclass[sigconf]{acmart}
\copyrightyear{2026}
\acmYear{2026}
\setcopyright{cc}
\setcctype{by}
\acmConference[CAIS '26]{ACM Conference on AI and Agentic Systems}{May 26--29, 2026}{San Jose, CA, USA}
\acmBooktitle{ACM Conference on AI and Agentic Systems (CAIS '26), May 26--29, 2026, San Jose, CA, USA}
\acmDOI{10.1145/3786335.3813126}
\acmISBN{979-8-4007-2415-2/2026/05}
\usepackage{graphicx}
\usepackage{xcolor}
\usepackage{csquotes}
\usepackage{listings}
\usepackage{subcaption}

\definecolor{mygreen}{rgb}{0,0.6,0}
\definecolor{mygray}{rgb}{0.5,0.5,0.5}
\definecolor{mymauve}{rgb}{0.58,0,0.82}
\definecolor{darkgreen}{RGB}{0,128,0}
\definecolor{codebrown}{RGB}{150,100,0}
\lstdefinelanguage{ACDL}{
  keywords = [1]{@T, @t, @t.i, @1, sys, env, resp},
  keywords = [2]{S, U, T, A, N},
  keywords = [3]{Name, If, Else, ForEach},
 keywordstyle=\color{blue}\bfseries,
 keywordstyle=[2]\color{orange}\bfseries,
 keywordstyle=[3]\color{gray}\bfseries,
 literate=%
        {(}{{\textcolor{darkgreen}{(}}}{1}
        {)}{{\textcolor{darkgreen}{)}}}{1}
        {[}{{\textcolor{purple}{[}}}{1}
        {]}{{\textcolor{purple}{]}}}{1}
        {:}{{\textcolor{darkgreen}{:}}}{1}
        {\{}{{\textcolor{gray}{\{}}}{1}
        {\}}{{\textcolor{gray}{\}}}}{1}
}

\newcommand{\acdlfa}[2]{
\vspace{2pt}
\begin{minipage}{0.45\columnwidth}
  \lstinputlisting[language=ACDL,linewidth=0.95\textwidth]{acdl/#1.acdl}
\end{minipage}
\begin{minipage}{0.55\columnwidth}
\vspace{#2}
  \hfill
  \includegraphics[width=0.9\textwidth]{Figures/#1.png}\hspace{2pt}
\end{minipage}
}

\newcommand{\acdlf}[1]{
\acdlfa{#1}{0em}
}

\newcommand{\acdlv}[2]{
  \vspace{4pt}
  \includegraphics[width=#2\columnwidth]{Figures/#1.png}
}

\newcommand{\acdl}[1]{
  \begin{lstlisting}[language=ACDL]
  #1
  \end{lstlisting}
}

\newcommand{\y}[1]{\texttt{#1}}

\title{A Language for Describing Agentic LLM Contexts}

\date{February 27, 2026}

\begin{document}
\author{Noga Peleg Pelc}
\affiliation{
  \department{Dept. of Computer Science and AI}
  \institution{Bar-Ilan University}
  \country{}
}
\email{pelegpelcnoga@gmail}

\author{Gal A. Kaminka}
\affiliation{
  \department{Dept. of Computer Science and AI}
  \institution{Bar-Ilan University}
  \country{}
}
\email{galk@cs.biu.ac.il}

\author{Yoav Goldberg}
\affiliation{
  \department{Dept. of Computer Science and AI}
  \institution{Bar-Ilan University \& Ai2}
  \country{}
}
\email{yoav.goldberg@gmail}

\sloppy
\begin{abstract}
Large language models are increasingly used within larger systems (\enquote{LLM agents}). These make a sequence of LLM calls, each call providing the LLM with a combination of instructions, observations, and interaction history. The design of the encoded information and its structure play a central role in the quality of the resulting system, leading to efforts spent on context engineering. It is therefore critical to communicate the composition of the LLM context in a system, and how it evolves over time. Yet, no standard exists for doing so: context construction is typically conveyed through informal prose, ad hoc diagrams, or direct inspection of code, none of which precisely capture how a prompt evolves across interaction steps or how two context representation strategies differ. To remedy this, we introduce the Agentic Context Description Language (ACDL), a language for specifying the structure and dynamics of LLM input contexts in a precise, readable, and standard manner, along with visualizations. ACDL provides constructs for specifying context aspects such as role message sequences, dynamic content, time-indexed references, and conditional or iterative structure, capturing the full architecture of a prompt independently of any particular implementation. ACDL diagrams can be hand drawn on a whiteboard, or written in formal language which can then be rendered. We describe the language, demonstrate it by documenting several existing systems and their variants, and encourage the community to adopt it for describing LLM systems context, both in day-to-day communication and in papers. Tooling, examples and documentation are available at \url{www.acdlang.org}.
\end{abstract}

\maketitle

\section{Introduction}
Large language models are increasingly serving as the core reasoning component in complex, multi-component systems. Modern agentic systems, ranging from autonomous coders to multi-step reasoning frameworks, rely on sophisticated input contexts that evolve dynamically over time. The evolving structure of these contexts, including which messages are included, how history is accumulated, and under what conditions content appears, is a central design artifact that directly shapes agent behavior.

Despite the central role of context structure, there is currently no standard way to describe it, leaving practitioners to rely on informal prose, ad hoc diagrams, or direct inspection of implementation code. At the community level, the absence of a shared descriptive framework makes it challenging to compare context construction strategies across systems, especially when they appear similar at a high level. In addition, although many published systems provide accompanying code and prompt templates, the logic governing context assembly is rarely made explicit in the paper itself, complicating efforts to reproduce or meaningfully re-implement proposed methods. Finally, Within research and engineering teams, it becomes difficult to rigorously communicate how the context of an agent evolves over time or to reason about the impact of incremental changes in that context.

To address these challenges, we present \textbf{Agentic Context Description Language} (ACDL), a formal language for precisely describing the structure and evolution of context in multi-turn interactions between agents and large language models. ACDL abstracts away concrete content, replacing it with symbolic labels that describe the \emph{role}, \emph{type}, and \emph{source} of each element—distinguishing, for example, system instructions from user queries, or tool outputs from model reasoning. Temporal indices track how elements accumulate across interaction steps, while control flow constructs such as loops and marked regions capture the patterns by which context is assembled and transformed. This abstraction enables crisp communication and rigorous analysis of context structure and dynamics, independently of any particular implementation.

\begin{figure*}[t!]
    \centering
    \raisebox{0pt}[\height]{\begin{minipage}[t]{0.33\textwidth}
        \vspace{0pt}
        \includegraphics[width=0.9\textwidth]{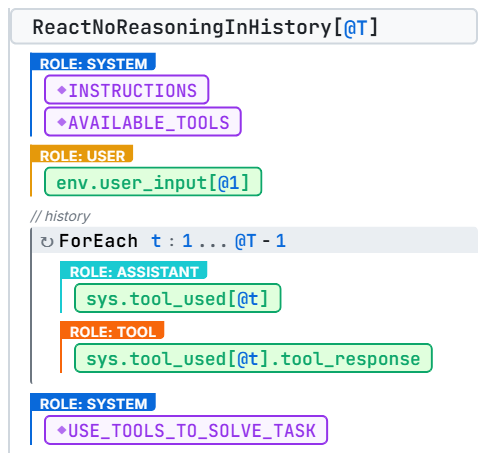}
    \end{minipage}}
    \hfill
    \raisebox{0pt}[\height]{\begin{minipage}[t]{0.33\textwidth}
        \vspace{0pt}
        \includegraphics[width=0.9\textwidth]{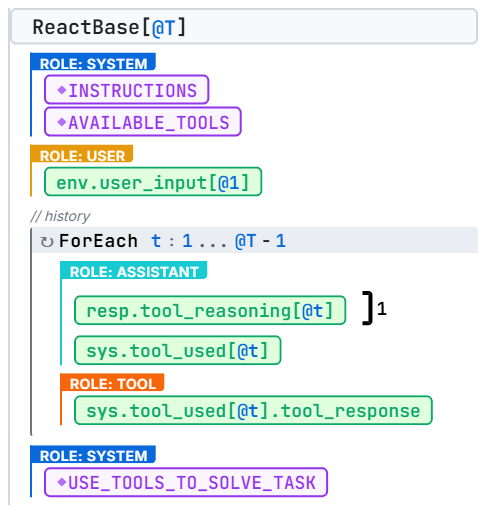}
    \end{minipage}}
    \hfill
    \raisebox{0pt}[\height]{\begin{minipage}[t]{0.33\textwidth}
        \vspace{0pt}
        \includegraphics[width=0.9\textwidth]{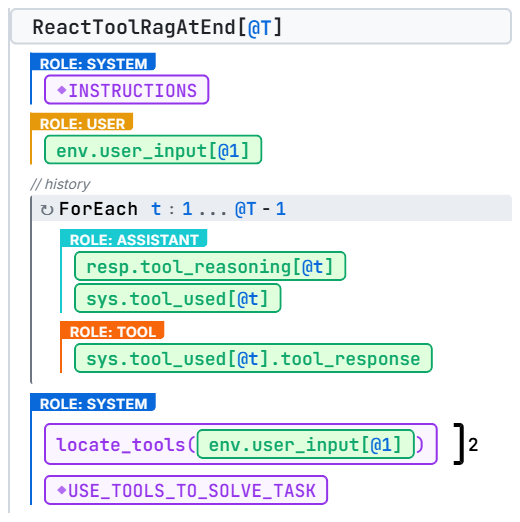}
    \end{minipage}}
    \caption{ACDL visualization of three simple ReAct loop variants. Middle: base implementation. Left: no reasoning traces in action history (without [1]). Right: query-based tool selection, tools appear at last rather than first message ([2]). ACDL makes expressing and discussing such differences concise, precise, and immediately apparent.}
    \label{fig:react-variants}
\end{figure*}

The value of ACDL is illustrated in Figure~\ref{fig:react-variants}, which presents three structurally distinct implementations of a multi-step ReAct\footnote{ReAct (Reasoning + Acting) is a prompting framework where a language model interleaves chain-of-thought reasoning with actions (such as calling tools or APIs) and observations from those actions, looping through thought, action, and observation steps until it reaches an answer.}\cite{react} agent rendered from their ACDL specification. All three implementations realize the same high-level design: a system prompt, user query, and iterative tool-use loop. However, they differ in where instructions appear, how assistant reasoning is interleaved with tool responses, and whether certain elements are repeated at each iteration. These variations reflect valid design choices that affect agent behavior, yet describing them in text does not easily reveal the differences between them (as we discuss in Section \ref{sec:eval}, even such small structural variations between ReAct implementations can lead to measurable differences in agent performance). ACDL renders each structural difference explicit, and the renderings are immediately comparable as differences in block locations and message roles are visible at a glance. Without a formal representation, articulating these distinctions, let alone systematically, would require lengthy prose or direct inspection of the code.

ACDL was designed to provide a shared language for describing and communicating about context structure in agentic LLM systems, to enable clearer communication within teams, more faithful comparison between published systems, and more transparent documentation of prompt logic in research articles and white-papers. It also has the potential for being a clear \emph{human-to-AI} communication medium, describing the desired context layout to coding agents in unambiguous terms. In addition to the language specification, we provide a parser, an interactive renderer, a vscode plugin, and an agentic skill that operationalize ACDL.

\section{Preliminaries: Agentic Systems Terminology}
We begin by clarifying the conceptual foundations underlying agentic systems, and the core challenges involved in building them. An \emph{agentic system} is an autonomous entity that operates in an iterative cycle of observing its environment, reasoning and making decisions about the appropriate course of action pursuing its own tasks, and taking actions according to these decisions. The decision-making process is sequential: later decisions take into account the current state as well as previous decisions and their results.
An agentic system operates its own time steps, maintaining its own memory which it uses to store and later retrieve information, to inform future behavior.
A multi-agent system comprises multiple agentic systems operating within a shared environment. Agentic systems within a multi-agent system may communicate with or observe the outputs of other agentic systems, and, in some instances, may share memory between agents. 
Together, these properties define the abstractions that the remainder of this paper builds upon.

\vspace{6pt}\noindent\textbf{LLM-based Agentic Systems. } 
In an \emph{LLM-based agentic system} some or all the decisions are delegated to an LLM.
At each decision point, the system specifies a situation, a set of possible parameterized actions to be taken, and supporting information. This information is then packed into a prompt and sent to the LLM (potentially together with additional instructions).
The information specified in the prompt is called \emph{the context} (see \emph{What's in a Context?} below).
The system interprets the LLM's response and acts according to it.
As LLM calls are independent from each other (stateless) yet the agent's decision making is ideally sequential (statefull), the context must contain enough information about the system's state and process history for the LLM to make the correct responses.

\vspace{6pt}\noindent\textbf{What's in a Context?}
Given the role of the context to communicate stateful information from one LLM invocation to the next, the context often includes \emph{a history} component, listing the past turns of the interaction and their outcomes.
In addition, the agentic system  may provide the LLM with general purpose instructions, information about the state of the environment, information about results of actions, information retrieved from long term memory and a set of possible actions to take (or tools to call). It may also list internal state information such as current goals and subgoals, plans, progress tracking towards achieving each of the subgoals or plans, reminders, etc. Just as history accumulates and therefore changes from one invocation to the next, so do other information components included in the context.

Changes may occur because of the system's actions that affect the environment, system actions that affect internal states, or changes to the environment that are independent of the system. It is convenient to talk about \enquote{steps} that the system takes, and talk about the context sent to the LLM at a specific step $t$. All the state values discussed above are also indexed according to these steps.

\vspace{6pt}\noindent\textbf{Describing Contexts vs Prompts. } 
We distinguish \emph{context engineering} from the closely-related, yet distinct, \emph{prompt engineering}. At some technical level, the terms \emph{context} and \emph{prompt} are interchangeable, and describe the entirety of  information sent as input to the LLM in a single call.
However, the terms are also used to distinguish different engineering concerns.

We use \emph{context engineering} to explicitly and purposefully abstract away from specific word choices and similar details that are left for \emph{prompt engineering}. Specifically, we use \emph{prompt engineering} to refer to tuning the specific wording of instructions, persona adoption, and phrasing used to nudge an LLM toward a desired output.
In contrast, we use \emph{context engineering} when discussing the macro-logic of information selection and structuring: what information is presented to the LLM at each stage, how it is packed into role messages, and where it is situated within the input.
Context engineering also addresses labeling of tool observations, how historical turns are prioritized, and how system instructions are protected from attention dilution.

The details of the context composition have a significant impact on the LLM system's performance. The following example illustrates how such a description may be carried out today.

\vspace{6pt}\noindent\textbf{Example Descriptions of LLM-Based Agentic Systems. }
A simple example of an LLM-based agentic system is a chat loop\footnote{We take a broad view of agentic systems, which also includes chat-bots.} where at each turn of the conversation the controller presents the LLM with all previous conversation turns (or a summary of them) followed by the latest user query.
Another example is a ReAct\cite{react} tool-calling loop, in which the system presents the LLM with a set of possible actions (\enquote{tools}), asks it to reason and then select the best one, applies the action, and repeats the process until a DONE action is selected. Here, the iteration is over action-selection steps, and at each step the LLM is presented with a summary of previous tool invocations and their outcomes.
A chat-bot and ReAct tool-calling scenarios can be combined into a system that converses with a user and, upon each user turn, runs an internal ReAct loop for producing a response, which is then sent back to the user. Here, the history sent to the LLM at each turn will comprise of the previous user requests, the previous system's responses to these requests, and some or all of the intermediary tool-calling requests and responses that resulted in these system responses.

The language used in the description above has been kept quite loose: it is not clear exactly how the history is structured and what information it contains or does not contain. While many papers denote more attention to refining such descriptions and making them clearer, it is in fact quite challenging. It is often the case that despite the best attempts of researchers and practitioners to clearly describe contexts in text or code, the description remains subject to interpretation.  This is the motivation for the our work.

\section{The Need for a Context Description Language}
Clear communication of design choices is a prerequisite for reproducible and improvable engineering. In computer science and software engineering, commonly used visual languages expressing system design include UML class diagrams, flowcharts, sequence diagrams, state diagrams, and reactive streams marble diagrams, to name a few, with software like Mermaid \cite{mermaid} allowing to express such diagrams in text-based markup.

In Machine Learning, \emph{plate-notation} expressing bayesian graphical models and \emph{factor-graph} notations were once ubiquitous and a major driving force for research advances in these fields. In deep-learning, the lack of a standard diagramming convention has been identified as a barrier to faithful implementation, comparison, and analysis of neural network architectures \cite{abbott2024, marshall2025}. The now-ubiquitous box diagrams used to describe architectures like ResNet or the Transformer---while informal and ad-hoc---have become an indispensable communication tool, enabling researchers to convey structural decisions and differences at a glance. Their value lies not in executability, but in making structure visible and comparable.

A well designed domain specific language (DSL) distills the core elements of the domain it attempts to communicate and their compositional patterns, and assigns them to a clear and consistent vocabulary. By so doing, it provides a shared vocabulary that enables practitioners to describe, compare, and reason about designs at the appropriate level of abstraction. Moreover, the existence of a language for describing artifacts has the potential of elevating these artifacts from after-thoughts to rigorous objects of study. Once a language exists to document structural differences, the variability space becomes apparent, and begs investigation.

In our view, context construction in agentic LLM systems is in dire need for such a language. There is currently no standard way to precisely describe how the input to a language model is assembled and how it evolves across interaction steps. For example, in a recent tech report, \citet{DeepSeek} describes how training data for chat conversations and training data for agentic tool use differ in the histories presented to the model. The verbal description is accompanied by two ad-hoc diagrams explaining the differences, and, while overall effective, both the text and diagrams do not fully capture the context structure. ACDL can be used to provide a unified and unambiguous description. Their text, diagrams and corresponding ACDL interpretations appear in Appendix \ref{deepseek}. 

While several languages have been proposed for structuring prompts or orchestrating LLM pipelines---including PromptML~\cite{promptml}, PDL~\cite{pdl}, and POML~\cite{poml}---these focus on organizing the content of individual prompts or on executing LLM-powered workflows, rather than on describing the mapping of temporally evolving system state and history to LLM contexts across multi-turn interactions.  

As we discuss in Section \ref{sec:eval}, even small variations in how a very basic ReAct agent's context is assembled---namely the inclusion or exclusion of reasoning steps from the history---lead to measurable differences in agent performance. These are precisely the kinds of distinctions that a descriptive language should make explicit. ACDL addresses this gap: it is not a programming language for building agents, but a descriptive language for specifying how context is composed, how it changes over time, and where its components originate.

\section{The Agentic Context Description Language}
\label{sec:acdl}

The Agentic Context Description Language (ACDL) is a a domain specific language for describing how LLM contexts are constructed as the state of the system evolves over time.
An ACDL specification describes the structure of the prompt presented to the model at each interaction step: which messages appear, in what order, with what content, and under what conditions. The language is descriptive---it specifies what the context contains, not how it is computed---and is independent of any particular codebase or runtime framework.
The language captures the structural blueprint of a prompt rather than its exact wording. It allows the description to refer to structural elements within past contexts or responses, and thus to describe how the context structure evolves.

\begin{figure}[h!]
    \centering
    \includegraphics[width=\columnwidth]{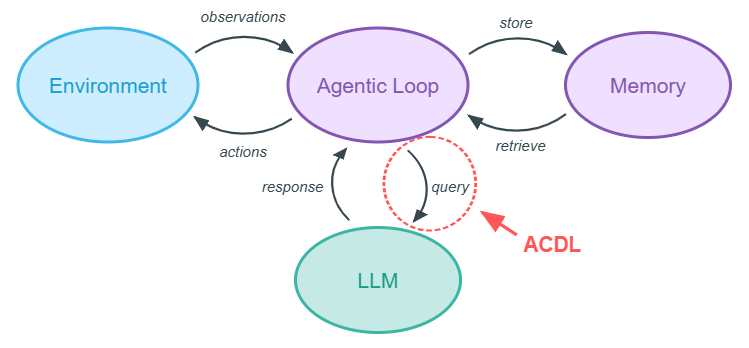}
    \caption{ACDL is concerned with describing the queries (contexts) sent to the LLM by an agentic controller.}
    \label{fig:env-diagram}
\end{figure}

\vspace{6pt}\noindent\textbf{Scope. } ACDL's scope is focused by design: precise description of the context window as seen by the language model (Figure \ref{fig:env-diagram}).
The context window sent to the LLM at time $T$ is derived from the state and the history at time $T$, and how they are translated to LLM contexts. ACDL describes this derivation logic.  It is \emph{not intended} to describe how state and history were produced, nor other aspects of the system:
agentic logic, tool implementations, retrieval mechanisms, model behavior and so on. These are explicitly out of scope for ACDL, and are meant to be described by other means.

\noindent ACDL is not meant to describe how state changes across steps, only how existing state and history are. are linearized into a context. A detailed description of ACDL syntax appears in Appendix~\ref{appendix:acdl}. Below, we use short examples to briefly describe major ACDL elements.

\vspace{6pt}
\noindent\textbf{Context as sequence of information pieces}
At the most abstract description, an LLM context is a linear sequence of \emph{information pieces}. In current LLM APIs, this sequence is packed into \emph{role messages} where each message is labeled with one of (currently) four distinct roles (\emph{user} (U), \emph{assistant} (A), \emph{system} (S), \emph{tool} (T)), and each message contains zero or more pieces of information. Thus, an ACDL context is a sequence of \emph{role messages} and each messages contains a sequence of information pieces.\footnote{Some new APIs break this assumption by supplementing the \texttt{messages} field with additional fields such as \texttt{tools}, whose content are then linearized into the context by the LLM provider, without user control on placement or formatting. Since these fields have relatively rigid structures, we currently leave them outside of ACDL, and expect their content to be specified in a comment or an accompanying description. Extending ACDL to support such fields is straightforward, and we may do so in the future.} Older LLM APIs (text completion) can be thought of as having a single role message with a role of \emph{none} (N).

The following is a minimal context described in ACDL, showing a simple sequence of two information pieces. The ACDL source is on the left; the visual rendition is on the right. The ACDL code defines a context named \texttt{Context}, applied at time \texttt{@T}. We will discuss time in more detail later. For now, it suffices to say that we assume the system operates in discrete time steps, counting integers from 1 onwards, and \texttt{@T} represents the current step. Time-step variables are marked with a @ prefix.

\acdlf{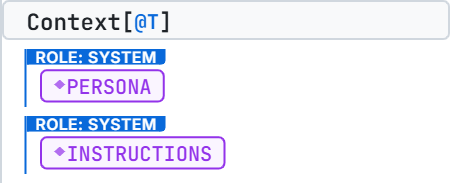}

The Context contains a sequence of two information pieces.
A single role message (\y{S}, indicating a \emph{system} role) with the content \y{PERSONA}, and a second \emph{system} message with the content \y{INSTRUCTIONS}.
Upper case symbols like \y{PERSONA} and \y{INSTRUCTIONS} represent constant strings (templates) that are defined elsewhere. While the context is indexed by time @T, these strings do not vary with time --- they are identical for all values of @T.
ACDL purposefully abstracts over the exact strings (which are a matter of prompt engineering), and focuses instead on the information pieces conveyed by these strings.

\vspace{6pt}\noindent\textbf{Information sources. }
While the pieces of information can be arbitrary strings, it is helpful to categorize them based on their source. These include:
(a) constant strings or templates; (b) values derived from the current system state; (c) values derived from the environment state; (d) values derived from previous LLM responses; and (e) values derived from functions applied to the above.

We now show how these elements manifest in ACDL.
One way is using templates that take arguments, which are pieces of information to be embedded in the string.
In the following example, the \y{INSTRUCTIONS} template takes two such arguments: a configuration-based "role" (``\emph{You are \{\{an expert coder\}\}}'') and the time at which the first step took place.

\acdlfa{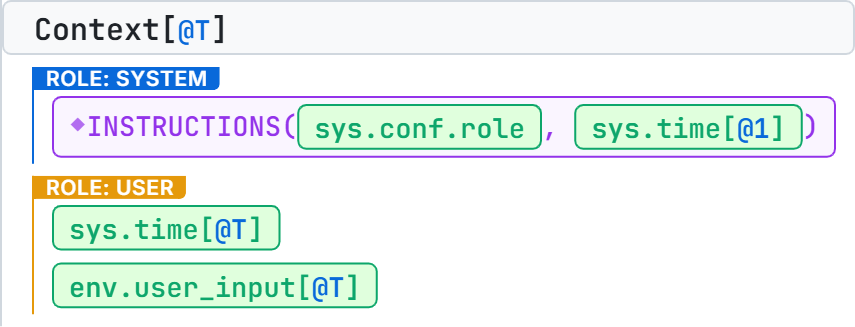}{2.5em}

\noindent It also adds a \emph{user} message with two pieces of information: the \emph{time} and the \emph{user input} at \emph{the current step}.
At each time step of the system, the LLM is presented with a fixed set of instructions in a system message, followed by the latest user input, timestamped with the time in which it occurred.
The user message likely contains some characters connecting the time and the user input. As ACDL focuses on the varying information and its source (rather than the exact strings), these are assumed to exist, but are not specified in ACDL.

The \emph{sys} and \emph{env} prefixes denote global objects that hold state information: \emph{sys} captures state maintained by the system (configuration, memory, action and tool-use history, etc.), while \emph{env} captures external state observed from the environment (user inputs, observed world state, etc.). Although both could share a single namespace, we keep them distinct because the distinciton between internal and external state is important for reasoning about system behavior and relates to issues such as trust, mutability, and persistence. An additional prefix is \emph{resp}, indicating an LLM response from a previous turn, whose value is used as is.

Every piece of information in the context is either constant, or derives from a state at a given time step.
ACDL does not define how the state is structured, and assumes the states and their history are maintained externally.
For example, accessing a property of \emph{sys}, \emph{env} or \emph{resp} assumes such a property is tracked by the system and contains the specified information.
Crucially, all values are immutable.
A value such as \emph{sys.conf.role} will remain the same throughout the lifetime of the system.
Values that change between states are accessed by indexing them with a state index: \emph{env.user\_input[@T]} is the value of the user input at the current step (@T).

The final kind of information source is \emph{functions}, demonstrated here in a simple RAG example (this example also shows assigning names to expressions with the \y{Name} keyword, and iteration over ranges and collections with the \y{ForEach} keyword, described later).
\noindent\acdlfa{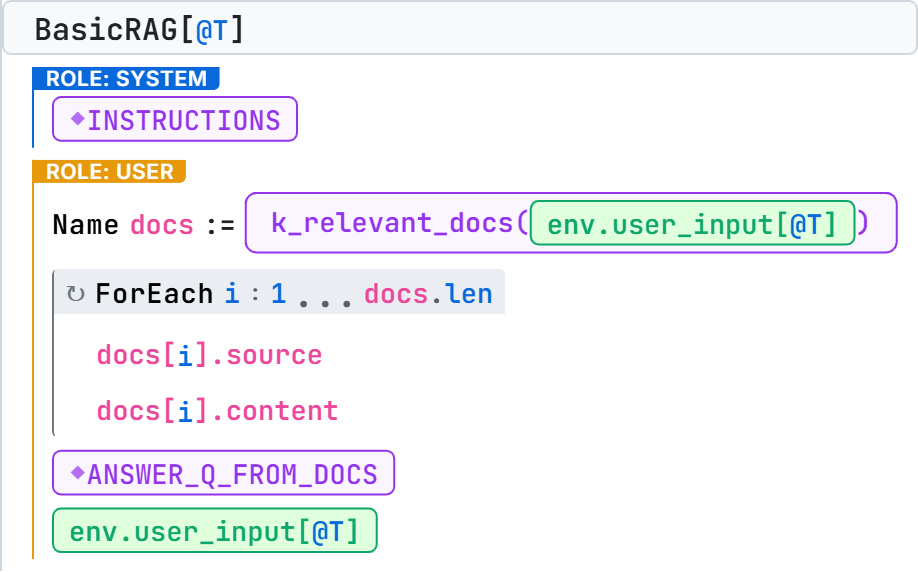}{7em}

\noindent \y{k\_relevant\_docs} is a function, computing values based on the parameters passed to it. Functions have descriptive names, and their semantics are either inferred by the reader or specified elsewhere (not in ACDL).
Here, the function returns a list of $k$ documents that are based on the current user's input. Each document contains (at least) a source and a content, which are included in the context.

Functions in ACDL are assumed to be pure, and their return values depend only on their parameters and external constants: calls to the same function with the same arguments always return the same results. If the function depends on state that changes through time (e.g., a memory that evolves as the system progresses) it should be reflected in their inputs: \emph{k\_relevant\_docs(env.user\_input[@T], sys.mem[@T], env.DB[@T])}.

\vspace{6pt}\noindent\textbf{Control Structures.}
In addition to iteration (see below), ACDL also supports conditions (\y{If}, \y{Else}, \y{Switch}). Control structures can appear either within a role message, as in the example above, or around role messages, as in the example below (a basic ReAct tool-calling agent):

\noindent\acdlv{react1}{0.5}

\noindent This examples demonstrates the most common use of ACDL \y{ForEach} iterations: unpacking history into the context. \y{React1} describes the context of a system that takes a user question as input, and then attempts to provide an answer by performing a series of tool calls, until it has sufficient information to answer the question reliably.

In the first time step, @1, the system sends the task description and available tools in a system prompt, followed by the user's input. In steps @2 and above, we enter the \y{ForEach} loop listing the history after the initial messages. The history is presented as alternating assistant and user messages, where the assistant messages list the requested tool and its reasoning at the given time in the history, and the user message contains the tool response.

Note that the very last step of the process, in which the LLM responds with the final answer which is returned to the user, is not part of the ACDL context: the system employing the \y{React1} context, will not send this information to the LLM, as it will produce the answer, finishing its job. ACDL describes what the LLM receives as context, not how the user experiences the system.

\vspace{6pt}\noindent\textbf{Time steps and sub-steps. } ACDL's model of agentic systems is of discrete-time state machines, operating on their own clock. Each state change in the system is a clock-tick (step increase), but a clock may tick also without a state change. Clock ticks often correlate with actions, as actions change state (hence moving the system to the next step). However, it is possible that several actions will result in a single step change: for example when the system collects all the action results internally before it updates its state based on all of their outcomes. The outside world may of course progress regardless of the system's clock. But observing the outside world amounts to a clock tick. The system's clock may progress due to: (a) external events from the environment (like the arrival of user input; (b) internal timers for periodic checking of state ("hearbeats"); or (c) the conclusion of some internal process that necessitates a state change (the conclusion of some actions or tool calls).

It is convenient to think of the system time-steps as being organized in a hierarchy. The top-level is what we consider to be the "main" steps. Major events that make the clock tick. This varies between system to system. For example, in a chat-bot system, user events trigger the system's operations, and hence the main time steps are driven by user inputs or their equivalent: each user input is a main step increase. In the React1 system above, there is only a single user input, which is then followed by a series of tool calls for obtaining an answer. Here, the main step is an iteration in the ReAct loop. But it is common for a system to work on both time scales. For example, \y{React2} below (left) describes a multi-turn, multi-step system, in which at each \emph{turn} (outer loop, chat-loop) a user enters a query and awaits a response, but producing each response involves a series of \emph{computation steps} (inner loop, ReAct loop).

This is handled by ACDL by introducing the concept of a sub-step, maintained by the system. The signature \y{React2[@T.I]} indicates the context at the $I$th substep of the $T$th main step.\footnote{Time steps can nest beyond two level, e.g., @T.I.J.K. Variable nesting levels are indicated by \texttt{@T.*}.} For each step \y{@t}, \y{@t.substeps} holds the number of sub-steps that participated in it. We can then access sub-steps via \y{@t.i} as done in the inner loop (2) and the final loop (3).\footnote{By convention, the current time step is indicated with upper letters, and the indices iterating over its range in the lower-cased version of the same letter.}

For nested loops, the context description may have multiple "exit points": here, at the main time steps \y{@T.0} the context concludes after the last user line, while for inner time-steps \y{@T.I > 0} the context concludes at the end. We can make this explicit with \y{PromptEndsHere} markers. Using them also helps to remove the trailing "last turn" in some cases (\y{React2Short}).

\noindent\begin{minipage}{0.5\columnwidth}
\acdlv{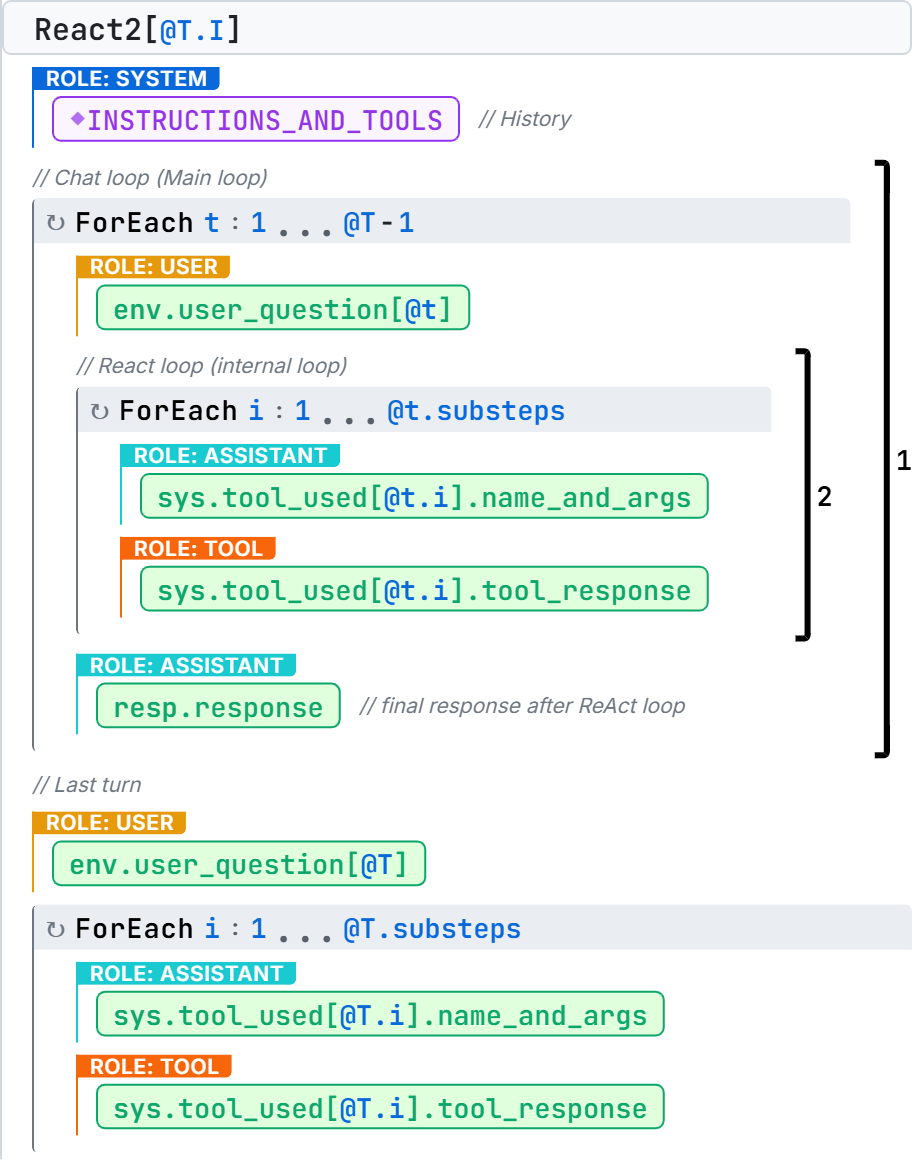}{1}
\end{minipage}
\begin{minipage}{0.5\columnwidth}
\acdlv{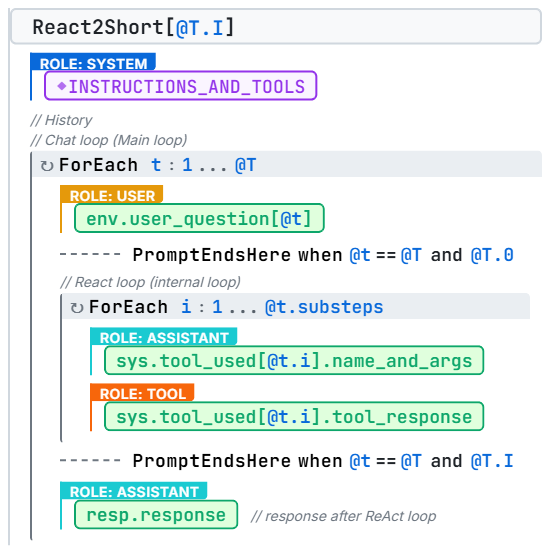}{1.0}
\vspace{4em} 
\end{minipage}

\noindent\textbf{Reuse through Fragments.} In some cases, the same context construction logic is used in multiple different LLM calls in the system, or even in different phases within the same LLM call. ACDL offers a mechanism called \emph{fragments} to capture this commonality and avoid repetition. A fragment receives parameters as input and results in either a string or a list of role messages. The semantics of invoking a fragment from within a context definition or another fragment is that the invocation site is replaced with the fragment value (invoking fragments resulting in strings are valid only within role messages, and invoking ones expanding to a list of role messages is valid only outside of them). The fragment definition syntax is very similar to context definition syntax. See details and examples in Appendix \ref{appendix:acdl}.

\vspace{6pt}
\noindent\textbf{Multi-agents.} Finally, ACDL can also represent contexts in multi-agent systems. In some agentic systems, e.g., as implemented in coding agents such as OpenCode, multiple agents are supported transparently: each agent is isolated from the others, having its own isolated state and clock, and ``agent invocation'' is handled as a tool calling. Every turn of the tool sub-agent is a tool-call for the invoker agent, and the sub-agent treats these calls as user inputs from the environment (where the user happens to be an agent). 

In contrast, some systems require a shared clock or shared memory, or that agents approach each other by name. To this end, the ACDL \y{Context} definition can be parameterized not only by the time step, but also by a variable indicating its agent id. Then, elements such as \y{sys[agent].foo} refer to a property specific to \y{agent} agent, and \y{sys[agent].foo[@t]} refers to this agent property at time @t. \y{sys.foo} and \y{sys.foo[@t]} refer to shared state. Agents who have access to another agent's ID, can use it within their own context, to refer to the other agent's state. In such systems, it is convenient if all agents share the same clock, and @t refers to the same clock. ACDL may be extended to support shared state without shared clocks in the future.

\begin{figure}
\centering
\includegraphics[width=0.85\columnwidth]{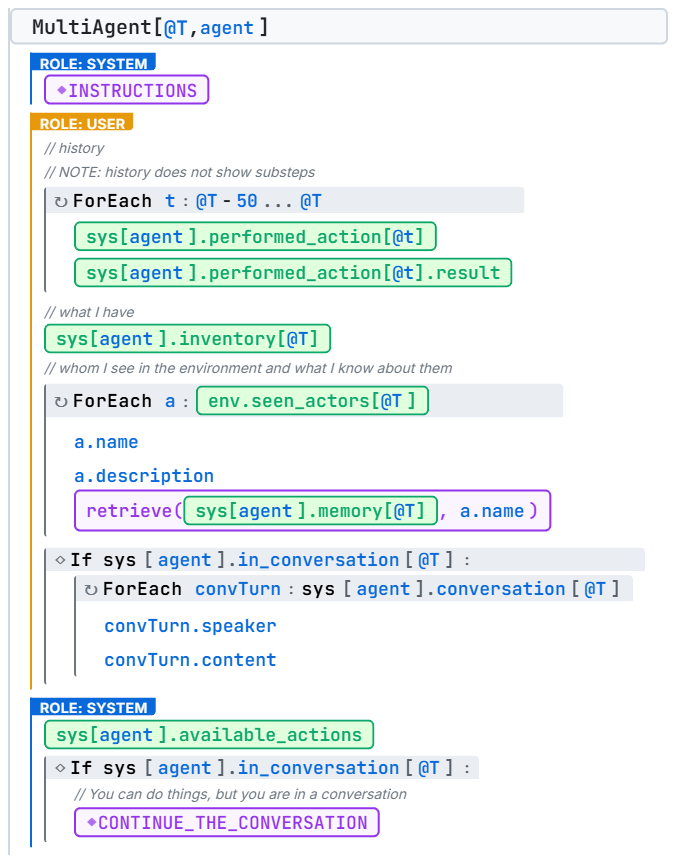}
\caption{ACDL Description of a simple multiagent main loop.}
\label{fig:multi-agent}
\end{figure}

Figure~\ref{fig:multi-agent} shows an example ACDL-defined context that includes an agent parameter. The context includes a history of the agent's last 50 actions and their results, and a description of the agent's current inventory. For each of the agent's peers observed in its environment, the context includes the peer's name, description, and details retrieved from memory. If the agent is currently engaged in a conversation with another agent, the context also includes a history of that conversation as a sequence of turns, each consisting of the speaker and what they said. Next, the agent is told what actions are available to it. Finally, if the agent is currently in a conversation, it is instructed to continue it.

\section{ACDL at Work}
\label{sec:eval}
The argument that a standard context description language would be helpful to the science and engineering of agentic systems is easy to make, as any standard language, if widely adopted, facilitates unambiguous communications, comparisons and analysis.
In this section, we attempt to make the more challenging argument, that \emph{ACDL is an excellent candidate standard for adoption}.

\vspace{6pt}
\subparagraph{\bf ACDL has tools to ease adoption.}
Tools facilitating the use of ACDL in describing contexts are made available (see \url{www.acdlang.org}).
We provide a web-based editor and visualizer that renders ACDL specifications and allows to download them as PDF or PNG; 
This is accompanied by complete language documentation (formatted for both humans and coding agents), as well as a set of example \texttt{.acdl} specifications that users can explore to familiarize themselves with the language and use as basis for their own. Additionally, we offer a VS Code extension with syntax highlighting and live rendering of ACDL files directly within the editor, along with a Claude Code skill \texttt{.md} file for an integrated authoring experience. The resources are open source, and we welcome bug fixes, extensions, and contributions. Finally, ACDL also lends itself to whiteboard diagramming in face-to-face communication.

Beyond its utility for human-to-human communication, ACDL may also serve as a medium for unambiguous human-to-AI communication. We anecdotally observed that Claude Code could read ACDL specifications and produce corresponding agentic loops, as well as convert from one agentic loop format to another. We flag this as a promising avenue for future work, and expect the underlying capability to improve rapidly as models progress.


\vspace{6pt}\subparagraph{\bf ACDL makes context nuances easy to locate and communicate.}
Through its visual rendering and the use of labels and numbered annotations, ACDL facilitates clear communications between those explaining the evolving structure of a context, and those trying to understand or compare it to others. 
From the point of view of a reader looking to understand the differences between contexts, they are apparent at a glance. 
From the point of view of an author, trying to communicate important details or differences, the use of numbered annotations in the visual rendering allows references from within the accompanying text. 

\begin{figure}
    \centering
    \includegraphics[width=0.9\columnwidth]{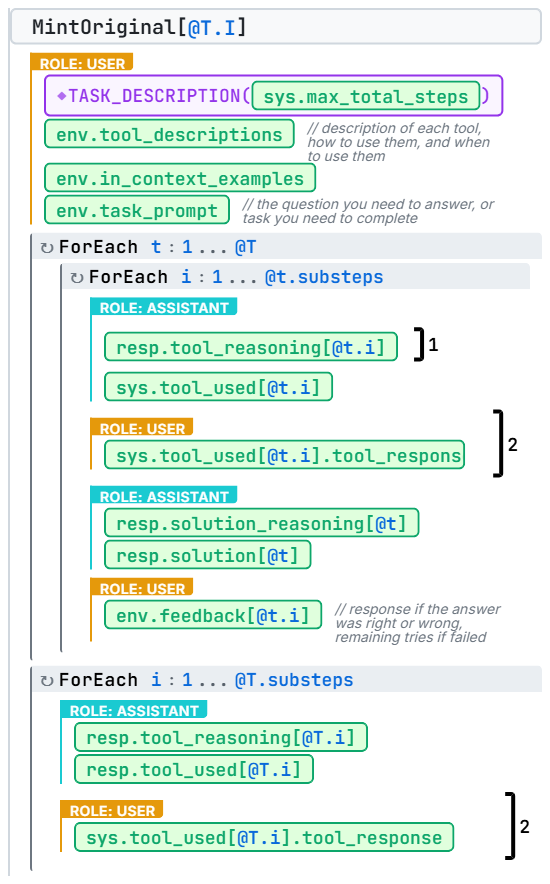}
    \caption{ACDL descriptions of a multi-turn multi-step ReAct agent as implemented in MINT \cite{mint}. The markings indicate variants: removing the line marked in (1); replacing the role of the line marked (2) from \emph{user} to \emph{tool}.}
    \label{fig:mint}
\end{figure}

For example, we consider the simple multi-turn multi-step ReAct agent as implemented in the MINT benchmark \cite{mint} and documented in ACDL in Figure \ref{fig:mint}. We then attempted a trivial variant: we remove the lines marked as (1)---removing reasoning traces from the tool-call history. This variant hurt performance on reasoning tasks while slightly improving performance on the code tasks. We tried 5 other variants in the experiment; see \ref{mint-appendix} for more details about this experiment.

\vspace{6pt}\subparagraph{\bf ACDL helps document and clarifies complex contexts.}
ACDL can go beyond simple examples, and scales to documenting elaborate contexts of game-playing systems, multi-agent systems and real-world state-of-the-art systems. We demonstrate it through examples: (1+2) documenting the main contexts of the popular OpenCode~\cite{opencode} 
and OpenClaw~\cite{openclaw} systems, as reverse-engineered by us from a combination of digging through code and inspecting proxy traces; and (3) documenting the Gemini Pokemon Blue agent as described meticulously in a detailed technical report \cite{Pokemon}. 

\begin{figure}[!t]
  \centering
  \includegraphics[width=\columnwidth]{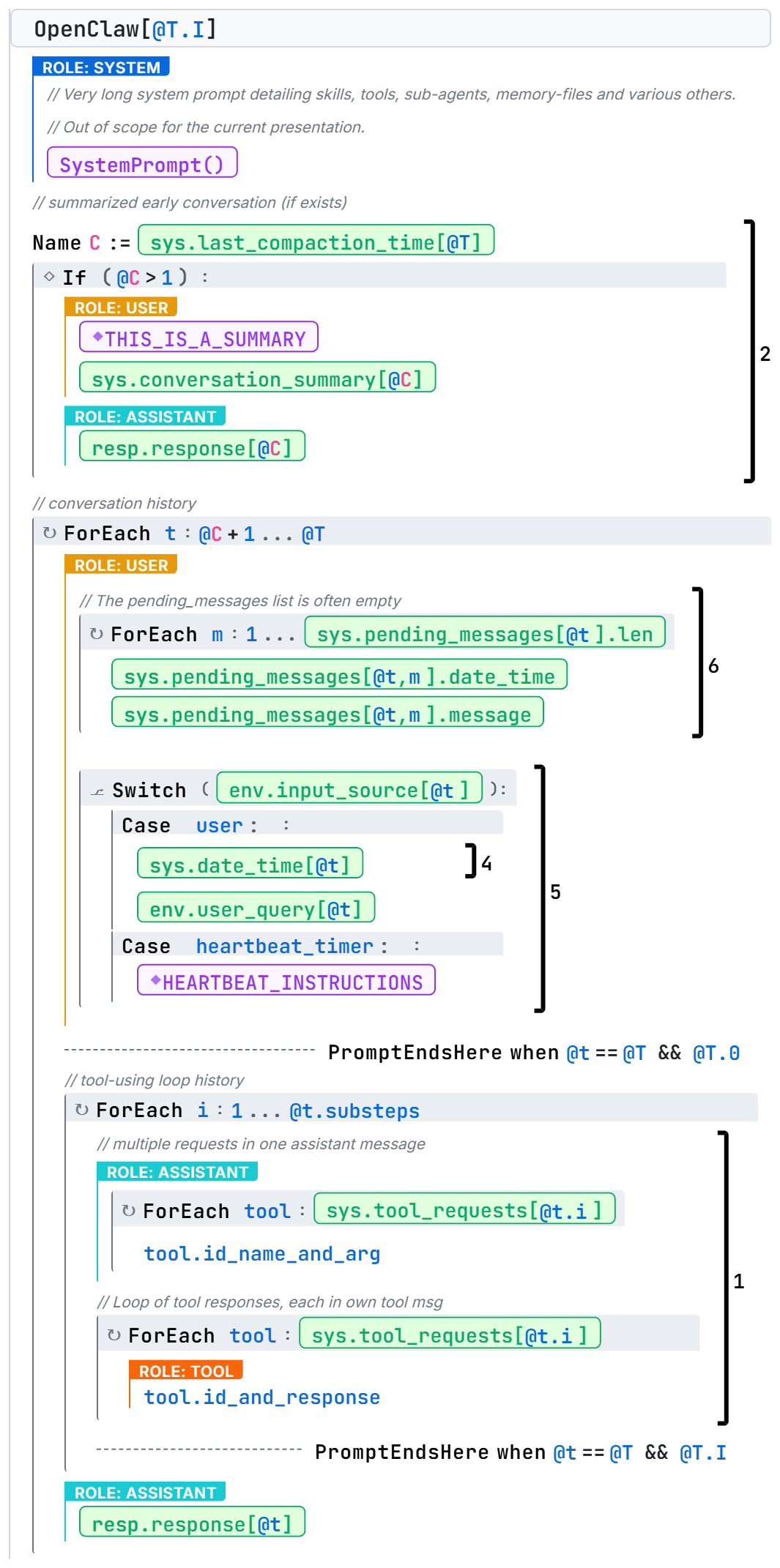}
  \caption{ACDL Description of OpenClaw main loop, without System Prompt details.}
  \label{fig:open-claw}
\end{figure}

\begin{figure}[!t]
  \centering
  \includegraphics[width=0.9\columnwidth]{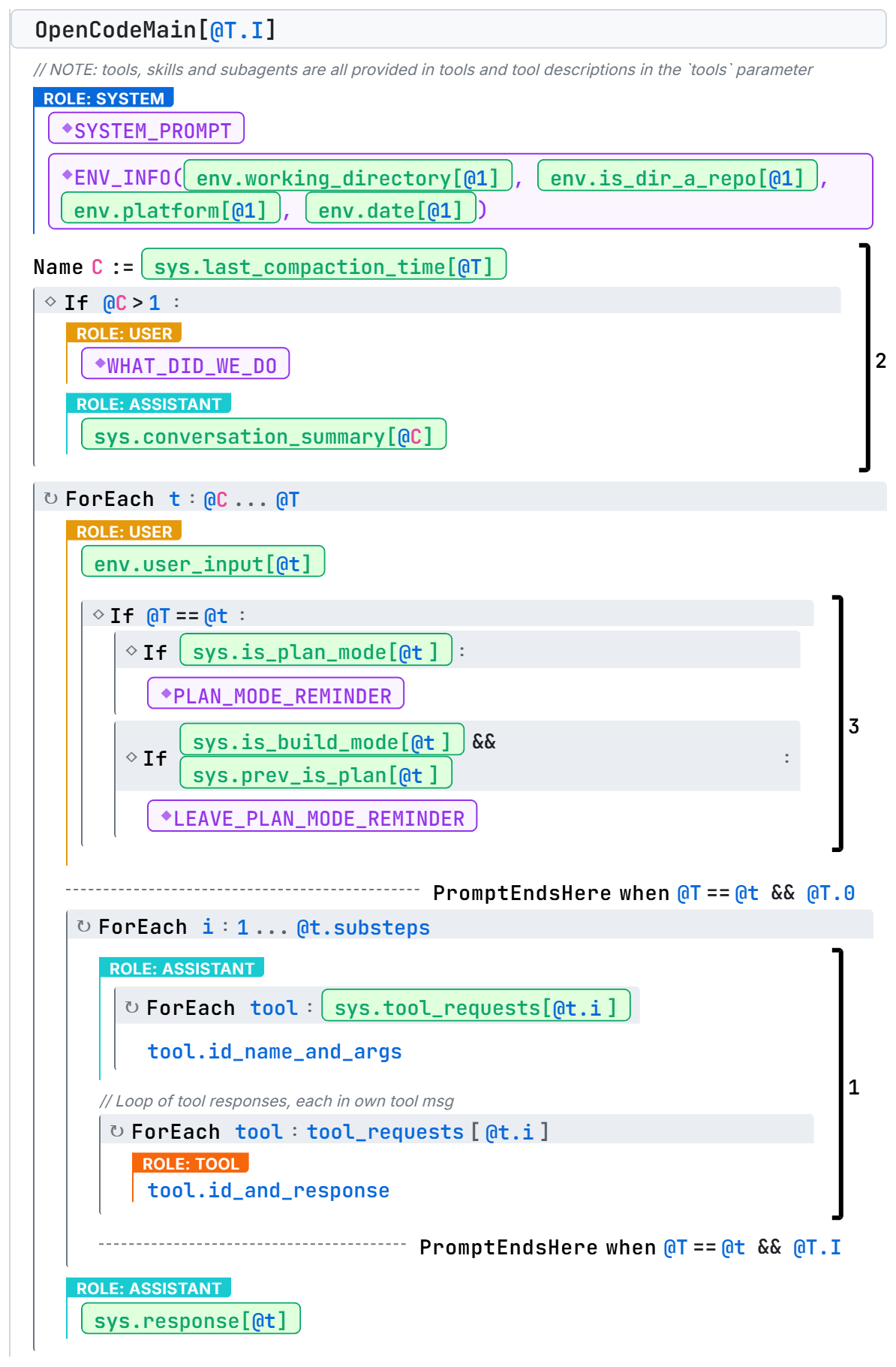}
  \caption{ACDL Description of OpenCode main loop. Tools, skills and sub-agents are passed in the \emph{tools} API parameter.}
  \label{fig:open-code}
\end{figure}

{\bf OpenCode and OpenClaw. }
OpenCode \cite{opencode} and OpenClaw \cite{openclaw} are two very popular open-source projects implementing widely-used agentic systems. OpenClaw is a personal assistant agent which can access the user's computer and performs tasks on their behalf, while OpenCode is a strong-performing coding-agent similar to claude-code. How are the contexts of these two systems implemented? We traced (reversed engineered) the main loop prompts of the two projects by looking at the code as well as at LLM-call traces, and documented them with ACDL (Figures \ref{fig:open-claw} and \ref{fig:open-code}).

Both systems follow the familiar multi-turn multi-step ReAct-loop pattern, where the lifetime of the loop is a "session" and most of the logic is implemented via tool-calling. In both systems, sub-agents are exposed as tools to be called by the main agent, with completely isolated contexts and states, hence are seamlessly supported. OpenClaw has a very long system prompt which includes also skills, tools, sub-agents and memories (not documented here). For OpenCode, the system prompt is more minimal, delegating skills, tools and sub-agent definitions to the "tools" parameter. The system prompt does include basic environment constants: working directory, git repository status, platform, and session start date.

Both systems also make use of the LLM ability to request several tool calls in the same turn. In both systems the tool-call requests for a step are appear in the history within a single assistant message, while their responses are presented in a sequence of individual tool messages (marked as (1)).

Both systems also support context compaction when the history grows too large. They differ somewhat in their summary presentation strategies (marked as 2): OpenClaw has an initial user message summarizing the previous conversation, followed by an assistant response to this summary. In contrast, OpenCode takes the opposite approach: an initial user message asks "what did we do so far" followed by an assistant message with the summary.

We also observe departures from and additions to the "classic" ReAct loop: OpenCode distinguishes between Plan-mode and Build-mode, and this is supported (marked as 3) by injecting "reminders" at the end of the last user message in the conversation.
OpenClaw includes a time-stamp in each user input (marked 4), supports non-user initiated tasks using a Hearbeat signal implemented as special user-messages (marked 5) and sometimes injects pending async messages on top of user-message content (6). ACDL makes these design similarities, as well as differences, apparent, inspectable, and amenable for discussion and potential cross-pollination and mutual improvements.

{\bf Gemini Plays Pok\'{e}mon Blue.} The Gemini 2.5 technical report~\cite{Pokemon} highlights the Gemini Plays Pok\'{e}mon project~\cite{zhang2025} as a demonstration of the model's agentic capabilities, in which Gemini 2.5 Pro was set up as an autonomous agent to play Pok\'{e}mon Blue from start to finish via a Twitch livestream, with the goal of beating the entire game (all 8 gym badges + Elite Four).

We include the verbatim description from the Gemini report in Appendix~\ref{app:pokemon}; here we provide a brief overview in our own words.
The agent receives observation information along with a system prompt explaining that it is playing Pok\'{e}mon Blue and that its goal is to beat the game. It is also given two tools (named \emph{Path Finder} and \emph{Boulder Puzzle Strategist}) and some general game-play tips. The agent then receives a summarized history of its previous actions: the most recent actions are included verbatim, while older history is compressed into summaries. Every 25 turns, a sub-agent evaluates progress on three main goals and several sub-goals, and the resulting progress report is injected into the context. Finally, the agent is instructed to explore and choose an action.

\begin{figure}[t]
    \centering
    \includegraphics[width=1\columnwidth]{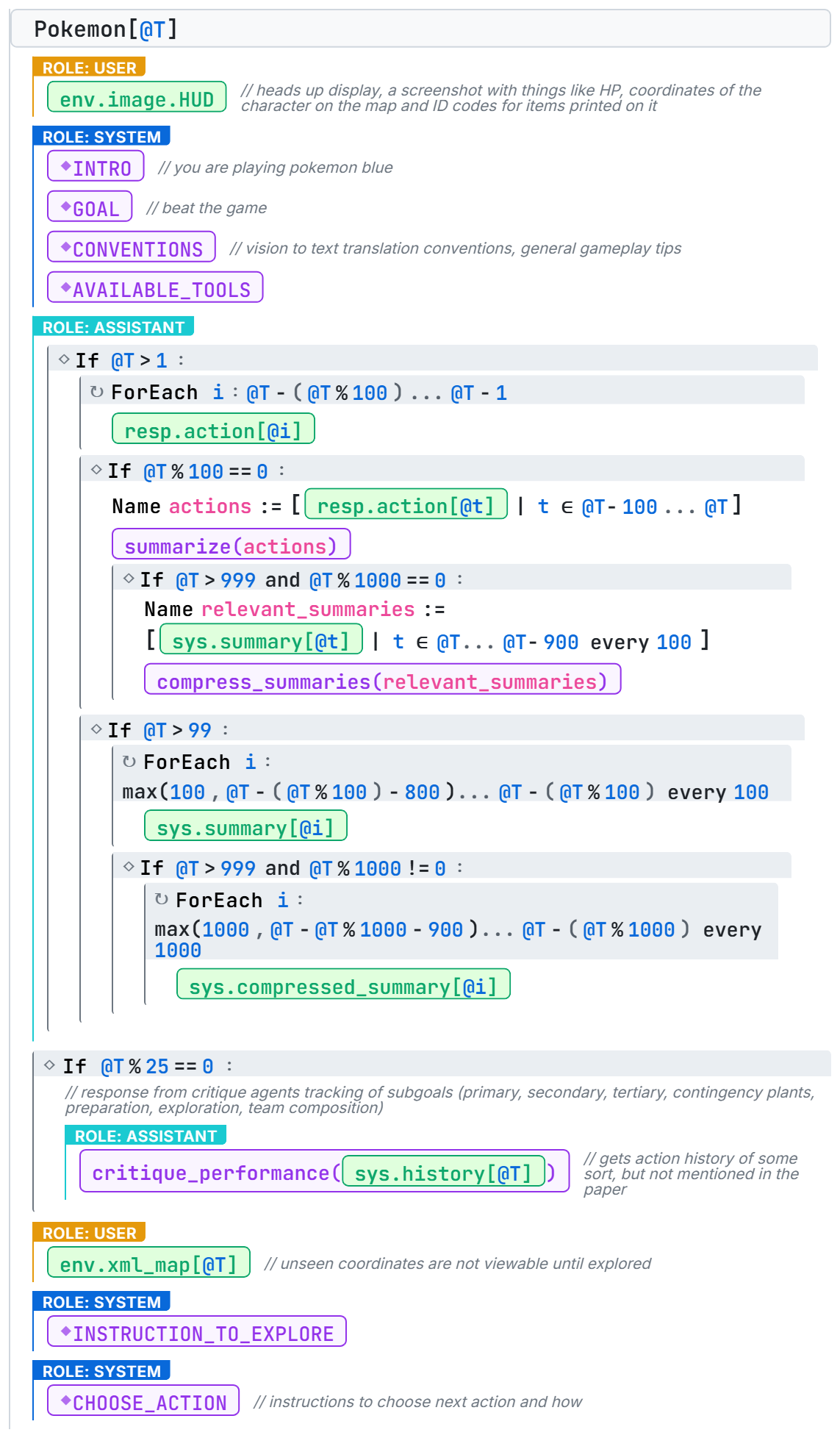}
    \caption{Gemini Plays Pok\'{e}mon Blue context structure visualization in ACDL.}
    \label{fig:Pokemon}
\end{figure}

The ACDL description of the Gemini report is much easier to understand and is less ambiguous, though it captures the same details as the report. The results in the visual rendition in Figure~\ref{fig:Pokemon}. It makes very clear what the history sent to the agent at each turn is, along with the data that it gets about the game and its goals. Additionally, the context specification makes use of a \textit{function} to show a call to a subagent that critiques the agent's decision by reviewing the history of his moves. 

\section{Related Work}
Several languages have been proposed to bring structure to prompt engineering, though their goals differ substantially from ACDL's.

PromptML \cite{promptml} is a domain-specific language for writing individual prompts as structured code. It provides explicit sections for context, objectives, instructions, examples, and metadata, giving prompt engineers a deterministic format for specifying what goes into a single prompt. PromptML focuses on the internal organization of one prompt rather than on how prompts evolve across interaction steps.

PDL (Prompt Declaration Language) \cite{pdl} is a YAML-based declarative language developed by IBM for composing LLM calls with tool use, code execution, and data processing into executable pipelines. PDL is a programming language for building LLM-powered applications: its specifications are programs that run, call models, and produce outputs. Its concern is the orchestration of computation, not the description of context structure.

POML (Prompt Orchestration Markup Language) \cite{poml}, developed by Microsoft, uses an HTML-like syntax with semantic tags such as \texttt{<role>}, \texttt{<task>}, and \texttt{<example>} to organize prompt components. It also provides data integration components for embedding documents, tables, and images, along with a CSS-like styling system for managing presentation. Like PromptML, POML is concerned with structuring the content of individual prompts rather than with describing how context changes over time.

ACDL differs from all three in both scope and intent. Where PromptML and POML focus on organizing the content within a single prompt, and PDL focuses on orchestrating executable LLM pipelines, ACDL is a language for describing how the full context is assembled and how its structure evolves across interaction steps. It is not designed to be executed or to produce prompts directly, but rather to make context construction strategies explicit, communicable, and comparable. Its core constructs—time indexing, control flow over interaction history, and cross-step references—address a concern that is largely absent from these other languages: the temporal dynamics of context in multi-turn agentic systems.

\section{Discussion}
The design of context structure and its evolution across multiple invocations is central to the performance of LLM-based agentic systems, yet the field lacks a standardized method for communicating the intricate  details needed for reproducing contexts, pointing out details, and communicating differences between contexts used in different systems.
To address this need, this paper introduced a candidate standard, the Agentic Context Description Language (ACDL). ACDL is a formal, human- and machine- readable, implementation-agnostic language that allows precise descriptions of complex context structures and their evolution across LLM invocations. ACDL descriptions are automatically rendered into visual descriptions.  

\paragraph{\textbf{Limitations.}}
We currently see two main limitations for ACDL. First, ACDL currently handles only agentic systems whose execution model is a sequence of (potentially nested) discrete time steps in which agent state mutates between LLM context constructions \emph{but remains immutable within each construction}. While technically all agentic systems \textit{can} be implemented this way, some real-world systems are not: they have mutable state that changes during context construction. Such systems are currently cumbersome to translate into ACDL descriptions, as they first have to be converted to an equivalent immutable-during-construction implementation and only then described. This friction is a real limitation, which we hope to mitigate in future versions.

Second, ACDL currently lacks a clean way to describe multi-agent systems in which the different agents have separate clocks (each agent runs according to its own time steps, without a synchronized clock), yet all have access to some \textit{shared} mutable state that can be read and modified by any of them. Such systems are plausible even if uncommon today. The only current workaround in ACDL is an explicit clock synchronization mechanism invoked when accessing the shared mutable state, which is inelegant.

We plan to address both of these in a future version, and will appreciate suggestions on how to do so.

\subsection*{Acknowledgments}
We gratefully acknowledge financial support from ISF Grant \#1373/24. As always, thanks to K. Ushi.

\bibliographystyle{ACM-Reference-Format}
\bibliography{references}


\begin{thebibliography}{13}


\ifx \showCODEN    \undefined \def \showCODEN     #1{\unskip}     \fi
\ifx \showISBNx    \undefined \def \showISBNx     #1{\unskip}     \fi
\ifx \showISBNxiii \undefined \def \showISBNxiii  #1{\unskip}     \fi
\ifx \showISSN     \undefined \def \showISSN      #1{\unskip}     \fi
\ifx \showLCCN     \undefined \def \showLCCN      #1{\unskip}     \fi
\ifx \shownote     \undefined \def \shownote      #1{#1}          \fi
\ifx \showarticletitle \undefined \def \showarticletitle #1{#1}   \fi
\ifx \showURL      \undefined \def \showURL       {\relax}        \fi
\providecommand\bibfield[2]{#2}
\providecommand\bibinfo[2]{#2}
\providecommand\natexlab[1]{#1}
\providecommand\showeprint[2][]{arXiv:#2}

\bibitem[Abbott(2024)]%
        {abbott2024}
\bibfield{author}{\bibinfo{person}{Vincent Abbott}.} \bibinfo{year}{2024}\natexlab{}.
\newblock \bibinfo{booktitle}{\emph{Neural Circuit Diagrams: Robust Diagrams for the Communication, Implementation, and Analysis of Deep Learning Architectures}}.
\newblock \bibinfo{type}{{T}echnical {R}eport} arXiv preprint arXiv:2402.05424. \bibinfo{institution}{arXiv}.
\newblock


\bibitem[Abdelaziz et~al\mbox{.}(2024)]%
        {pdl}
\bibfield{author}{\bibinfo{person}{Ibrahim Abdelaziz}, \bibinfo{person}{Kinjal Basu}, \bibinfo{person}{Mayank Agarwal}, \bibinfo{person}{Sadhana Kumaravel}, \bibinfo{person}{Matthew Stallone}, \bibinfo{person}{Rameswar Panda}, \bibinfo{person}{Yara Rizk}, {et~al\mbox{.}}} \bibinfo{year}{2024}\natexlab{}.
\newblock \bibinfo{booktitle}{\emph{{PDL}: A Declarative Prompt Programming Language}}.
\newblock \bibinfo{type}{{T}echnical {R}eport} arXiv preprint arXiv:2410.19135. \bibinfo{institution}{arXiv}.
\newblock
\urldef\tempurl%
\url{https://arxiv.org/abs/2410.19135}
\showURL{%
\tempurl}


\bibitem[{DeepSeek-AI}(2026)]%
        {DeepSeek}
\bibfield{author}{\bibinfo{person}{{DeepSeek-AI}}.} \bibinfo{year}{2026}\natexlab{}.
\newblock \bibinfo{title}{{DeepSeek-V4: Towards Highly Efficient Million-Token Context Intelligence}}.
\newblock \bibinfo{howpublished}{\url{https://huggingface.co/deepseek-ai/DeepSeek-V4-Pro/blob/main/DeepSeek_V4.pdf}}.
\newblock
\newblock
\shownote{Technical report}.


\bibitem[{Gemini Team, Google}(2025)]%
        {Pokemon}
\bibfield{author}{\bibinfo{person}{{Gemini Team, Google}}.} \bibinfo{year}{2025}\natexlab{}.
\newblock \bibinfo{booktitle}{\emph{Gemini 2.5: Pushing the Frontier with Advanced Reasoning, Multimodality, Long Context, and Next Generation Agentic Capabilities}}.
\newblock \bibinfo{type}{{T}echnical {R}eport}. \bibinfo{institution}{Google}.
\newblock
\urldef\tempurl%
\url{https://storage.googleapis.com/deepmind-media/gemini/gemini_v2_5_report.pdf}
\showURL{%
\tempurl}


\bibitem[Marshall et~al\mbox{.}(2025)]%
        {marshall2025}
\bibfield{author}{\bibinfo{person}{Guy Marshall}, \bibinfo{person}{Andr{\'e} Freitas}, {and} \bibinfo{person}{Caroline Jay}.} \bibinfo{year}{2025}\natexlab{}.
\newblock \showarticletitle{An Evidence-Based Guidance Framework for Neural Network System Diagrams}.
\newblock \bibinfo{journal}{\emph{PLOS ONE}} \bibinfo{volume}{20}, \bibinfo{number}{3} (\bibinfo{year}{2025}), \bibinfo{pages}{e0318800}.
\newblock
\href{https://doi.org/10.1371/journal.pone.0318800}{doi:\nolinkurl{10.1371/journal.pone.0318800}}


\bibitem[{OpenCode}(2025)]%
        {opencode}
{OpenCode} \bibinfo{year}{2025}\natexlab{}.
\newblock \bibinfo{title}{OpenCode: The Open Source {AI} Coding Agent}.
\newblock \bibinfo{howpublished}{\url{https://www.opencode.ai/}}.
\newblock
\newblock
\shownote{Open-source provider-agnostic coding agent for terminal, desktop, and IDE}.


\bibitem[Steinberger(2025)]%
        {openclaw}
\bibfield{author}{\bibinfo{person}{Peter Steinberger}.} \bibinfo{year}{2025}\natexlab{}.
\newblock \bibinfo{title}{OpenClaw: Your Own Personal {AI} Assistant}.
\newblock \bibinfo{howpublished}{\url{https://www.openclaw.ai/}}.
\newblock
\newblock
\shownote{Open-source autonomous AI agent framework. Originally released as Clawdbot, November 2025}.


\bibitem[Sveidqvist and {Contributors to Mermaid}(2014)]%
        {mermaid}
\bibfield{author}{\bibinfo{person}{Knut Sveidqvist} {and} \bibinfo{person}{{Contributors to Mermaid}}.} \bibinfo{year}{2014}\natexlab{}.
\newblock \bibinfo{title}{Mermaid: Generate diagrams from markdown-like text}.
\newblock
\urldef\tempurl%
\url{https://mermaid.ai/}
\showURL{%
\tempurl}


\bibitem[Wang et~al\mbox{.}(2024)]%
        {mint}
\bibfield{author}{\bibinfo{person}{Xingyao Wang}, \bibinfo{person}{Zihan Wang}, \bibinfo{person}{Jiateng Liu}, \bibinfo{person}{Yangyi Chen}, \bibinfo{person}{Lifan Yuan}, \bibinfo{person}{Hao Peng}, {and} \bibinfo{person}{Heng Ji}.} \bibinfo{year}{2024}\natexlab{}.
\newblock \showarticletitle{MINT: Evaluating LLMs in Multi-turn Interaction with Tools and Language Feedback}. In \bibinfo{booktitle}{\emph{Proceedings of the Twelfth International Conference on Learning Representations (ICLR)}}.
\newblock
\urldef\tempurl%
\url{https://arxiv.org/abs/2309.10691}
\showURL{%
\tempurl}


\bibitem[Yao et~al\mbox{.}(2023)]%
        {react}
\bibfield{author}{\bibinfo{person}{Shunyu Yao}, \bibinfo{person}{Jeffrey Zhao}, \bibinfo{person}{Dian Yu}, \bibinfo{person}{Nan Du}, \bibinfo{person}{Izhak Shafran}, \bibinfo{person}{Karthik Narasimhan}, {and} \bibinfo{person}{Yuan Cao}.} \bibinfo{year}{2023}\natexlab{}.
\newblock \showarticletitle{{ReAct}: Synergizing Reasoning and Acting in Language Models}. In \bibinfo{booktitle}{\emph{International Conference on Learning Representations (ICLR)}}.
\newblock


\bibitem[Yellavula(2024)]%
        {promptml}
\bibfield{author}{\bibinfo{person}{Naren Yellavula}.} \bibinfo{year}{2024}\natexlab{}.
\newblock \bibinfo{title}{{PromptML}: Prompt Markup Language}.
\newblock
\urldef\tempurl%
\url{https://www.promptml.org}
\showURL{%
\tempurl}
\newblock
\shownote{GitHub repository}.


\bibitem[Zhang(2025)]%
        {zhang2025}
\bibfield{author}{\bibinfo{person}{Joel Zhang}.} \bibinfo{year}{2025}\natexlab{}.
\newblock \bibinfo{title}{The Making of Gemini Plays Pok\'{e}mon}.
\newblock
\urldef\tempurl%
\url{https://blog.jcz.dev/the-making-of-gemini-plays-pokemon}
\showURL{%
\tempurl}
\newblock
\shownote{Blog post}.


\bibitem[Zhang et~al\mbox{.}(2025)]%
        {poml}
\bibfield{author}{\bibinfo{person}{Yuge Zhang}, \bibinfo{person}{Nan Chen}, \bibinfo{person}{Jiahang Xu}, {and} \bibinfo{person}{Yuqing Yang}.} \bibinfo{year}{2025}\natexlab{}.
\newblock \bibinfo{booktitle}{\emph{Prompt Orchestration Markup Language}}.
\newblock \bibinfo{type}{{T}echnical {R}eport} arXiv preprint arXiv:2508.13948. \bibinfo{institution}{arXiv}.
\newblock
\urldef\tempurl%
\url{https://arxiv.org/abs/2508.13948}
\showURL{%
\tempurl}


\end{thebibliography}

\clearpage
\section*{Appendices}
\appendix

\section{The MINT Experiment}
\label{mint-appendix}
We evaluated 7 context configurations that share the same system message structure---task description, tool descriptions, in-context examples, and task prompt---but differ along two structural axes: how much tool-use history is retained across turns, and how tool calls and their responses are grouped into messages.

\begin{figure*}
  \centering
  \begin{subfigure}[t]{0.30\textwidth}
    \vspace{0pt}
    \centering
    \includegraphics[width=\linewidth]{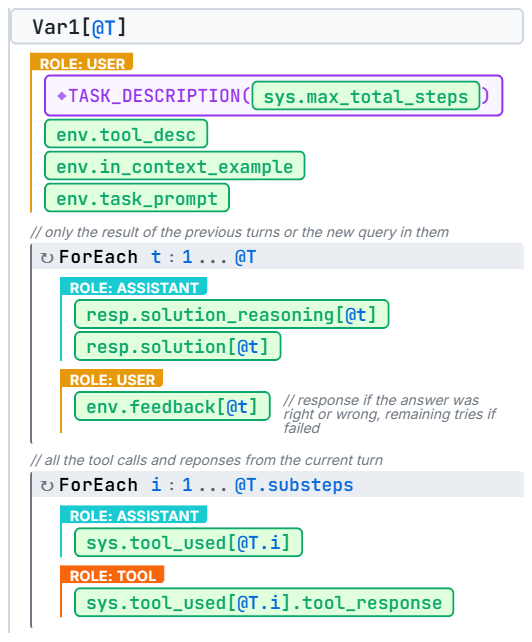}
    \caption{Variation 1. Previous turns include only the final reasoning and solution, omitting thinking and tool traces. The current turn includes the tool call in an assistant message and the tool response in a separate tool-role message.}
    \label{fig:mint-var1}
  \end{subfigure}\hfill
  \begin{subfigure}[t]{0.30\textwidth}
    \vspace{0pt}
    \centering
    \includegraphics[width=\linewidth]{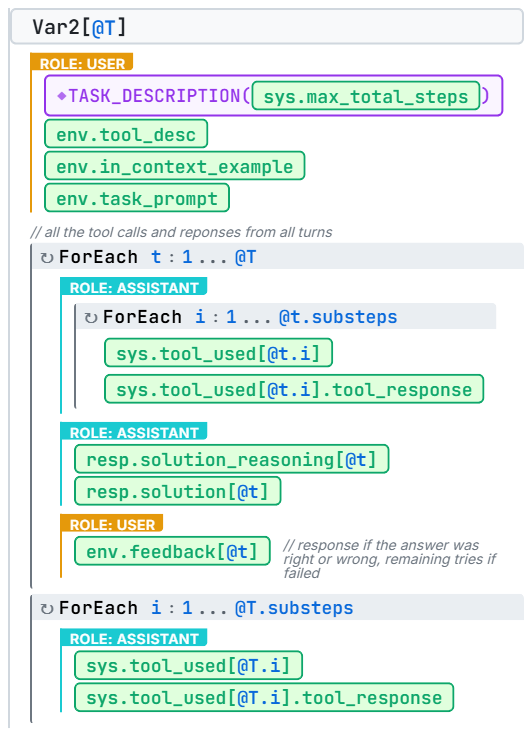}
    \caption{Variation 2. Previous turns include all tool calls and responses, with each turn collapsed into a single message. The current turn places each tool call and its response in their own messages.}
    \label{fig:mint-var2}
  \end{subfigure}\hfill
  \begin{subfigure}[t]{0.30\textwidth}
    \vspace{0pt}
    \centering
    \includegraphics[width=\linewidth]{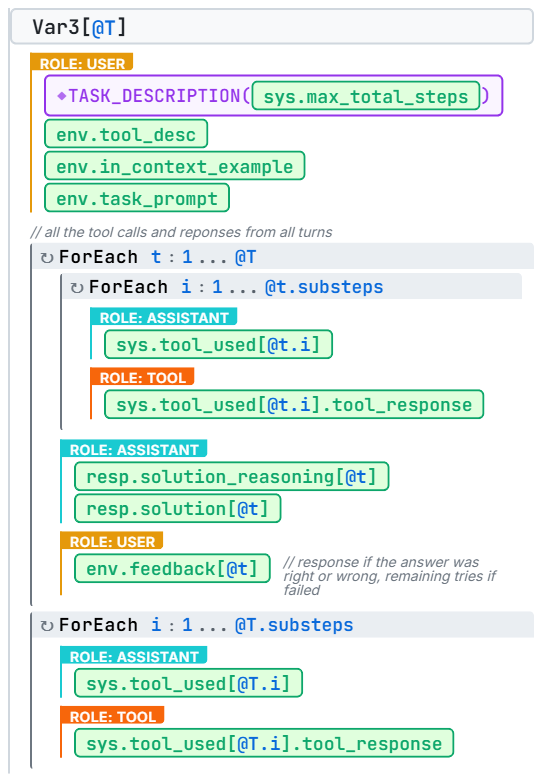}
    \caption{Variation 3. Previous turns include all tool calls and responses, with each tool call in a separate assistant message and each response in a separate tool-role message. The current turn is structured as in Variation 1.}
    \label{fig:mint-var3}
  \end{subfigure}
  \vspace{1em}
  \begin{subfigure}[t]{0.30\textwidth}
    \vspace{0pt}
    \centering
    \includegraphics[width=\linewidth]{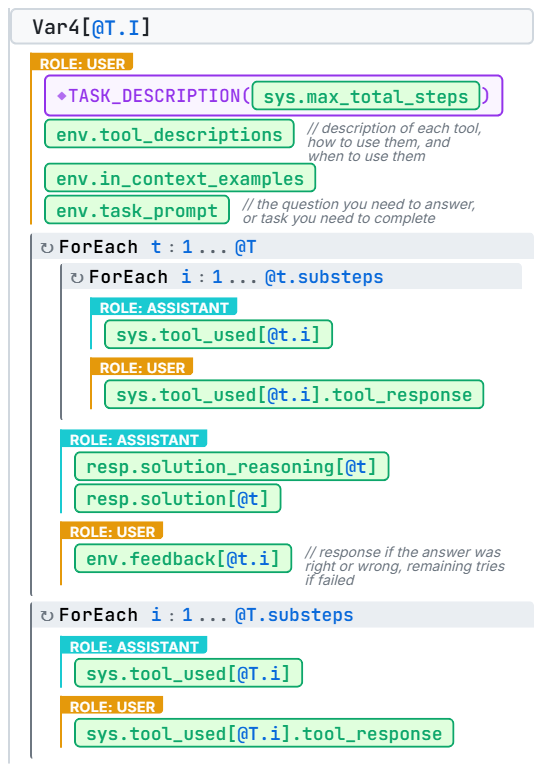}
    \caption{Variation 4. Same as Variation 3, but tool responses are placed in user-role messages instead of tool-role messages.}
    \label{fig:mint-var4}
  \end{subfigure}\hfill
  \begin{subfigure}[t]{0.30\textwidth}
    \vspace{0pt}
    \centering
    \includegraphics[width=\linewidth]{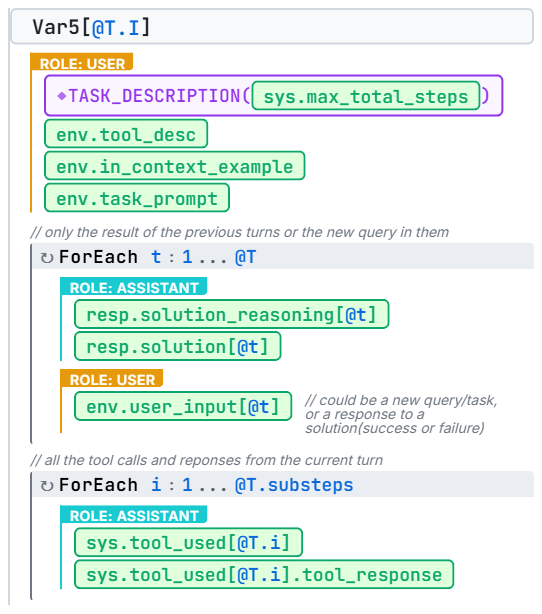}
    \caption{Variation 5. Same as Variation 1, but the final tool response is merged into the preceding assistant message instead of appearing in its own tool-role message.}
    \label{fig:mint-var5}
  \end{subfigure}\hfill
  \begin{subfigure}[t]{0.30\textwidth}
    \vspace{0pt}
    \centering
    \includegraphics[width=\linewidth]{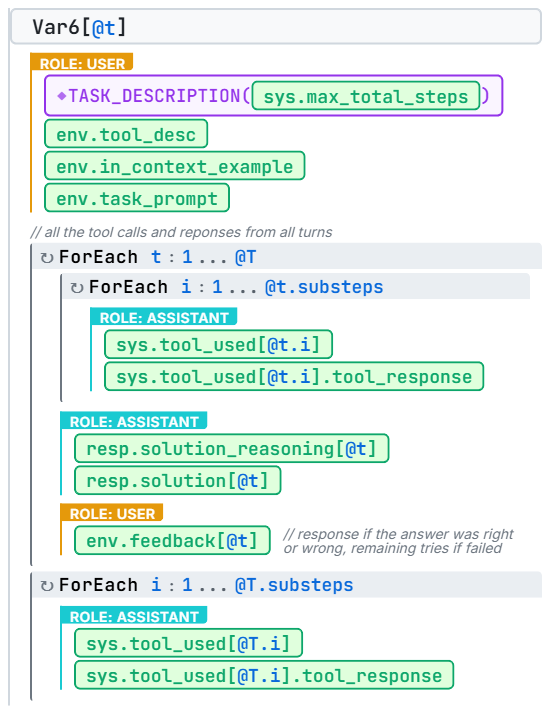}
    \caption{Variation 6. Previous turns split each tool interaction across two messages: an assistant message for the tool call and a tool-role message for the response. The current turn is structured as in Variation 2.}
    \label{fig:mint-var6}
  \end{subfigure}
  \caption{Six variations of the Mint context explored in our experiments. Across all variations, the user instruction message and the reporting of previous turns' solutions, reasoning, and feedback remain unchanged.}
  \label{fig:mint-variations}
\end{figure*}

We evaluate all configurations on the MINT benchmark tasks (excluding ALFWorld). The reasoning tasks are drawn from GSM8K, HotpotQA, MATH, MMLU, and TheoremQA, while the code-generation tasks are drawn from MBPP and HumanEval. We use GPT-4-Turbo as the underlying LLM. The results are summarized in the tables below.

\begin{table}[h]
\centering
\begin{subtable}[t]{0.48\textwidth}
\centering
\caption{Reasoning tasks}
\begin{tabular}{|l|c|}
\hline
\textbf{Variation} & \textbf{Success Rate} (\%) \\
\hline
var2          & 77.53 \\
\hline
mint-original & 76.90 \\
\hline
var6          & 76.90 \\
\hline
var1          & 74.68 \\
\hline
var4          & 74.37 \\
\hline
var5          & 74.05 \\
\hline
var3          & 72.78 \\
\hline
\end{tabular}
\end{subtable}
\vspace{0.5em}

\begin{subtable}[t]{0.48\textwidth}
\centering
\caption{Code generation tasks}
\begin{tabular}{|l|c|}
\hline
\textbf{Variation} & \textbf{Success Rate} (\%) \\
\hline
var6          & 54.41 \\
\hline
var3          & 54.41 \\
\hline
var4          & 54.41 \\
\hline
mint-original & 52.21 \\
\hline
var2          & 51.47 \\
\hline
var5          & 50.74 \\
\hline
var1          & 48.53 \\
\hline
\end{tabular}
\end{subtable}
\end{table}

These choices clearly affect performance: the gap between the best and worst configurations reaches nearly 5 percentage points even in this simple setup, the best variation for reasoning differs from that for code generation, and the MINT-original formulation is not the top performer in either category.

We used GPT-4-Turbo, a relatively weak model by current standards, because stronger models such as GPT-5 or claude-opus 4.6 are largely insensitive to these distinctions on MINT, where prompts are short, tasks are simple, and time horizons are limited. This insensitivity does not imply that context engineering no longer matters; rather, it reflects the fact that academic benchmarks with short interaction horizons fail to surface effects that would likely emerge in real-world, long-horizon agentic systems.

\newpage
\section{Gemini plays Pok\'{e}mon blue}
\label{app:pokemon}
The following is taken verbatim from the Gemini 2.5 technical report~\cite{Pokemon}, describing the Gemini Plays Pok\'{e}mon agent~\cite{zhang2025}:
\begin{quote}
The Gemini Plays Pok\'{e}mon agent \cite{zhang2025} receives a subset of RAM information, intended to give sufficient information to play the game, partially overlaid with a screenshot of the Game Boy screen. Gemini is prompted with a system prompt telling it that it is playing Pok\'{e}mon Blue and that its goal is to beat the game, as well as descriptive information to help it understand the conventions in the translation from vision to text and a small number of general tips for gameplay. Gemini then takes actions, translated to button presses. The sequence of actions is stored in context, followed by a summary clear every 100 turns. The summaries are stored in context as well. Every 1000 turns GPP compresses the existing summaries again. Additionally, Gemini keeps track of three main goals (primary, secondary, and tertiary) as well as several additional goals (contingency plans, preparation, exploration, team composition). Every 25 turns, another prompted instance of Gemini (Guidance Gemini, or GG) observes the same context as the main Gemini and critiques performance and attempts to point out hallucinations and so on. The overworld fog-of-war map is stored in the context in XML, where coordinates which have not been seen cannot be viewed until explored. Crucially, in the system prompt, Gemini is instructed to explore. Once a tile is explored, however, the coordinate is automatically stored in the map memory and labeled with a visited counter. Tiles are also labeled by type (water, ground, cuttable, grass, spinner, etc.), and warp points to different maps are also labeled as such. Gemini also has access to two agentic tools, which are both instances of Gemini equipped with a more specialized prompt---the \texttt{pathfinder} tool, and the \texttt{boulder\_puzzle\_strategist} tool. In the \texttt{pathfinder} prompt, Gemini is prompted to mentally simulate a path-finding algorithm, which is left unspecified, and to verify that the path is valid against the map information available. In the \texttt{boulder\_puzzle\_strategist} tool, Gemini is prompted to solve special boulder puzzles that are present in Pok\'{e}mon Blue in the Victory Road dungeon---these puzzles are similar to the game Sokoban---again, by mentally simulating sequences of actions that lead to solutions to the puzzle. The prompt describes the physics and the task of the boulder puzzle, as well as the desired output of solutions.
\end{quote}

\clearpage
\section{DeepSeek-V4}
\label{deepseek}
\begin{figure}[H]
  \centering
  \includegraphics[width=\columnwidth]{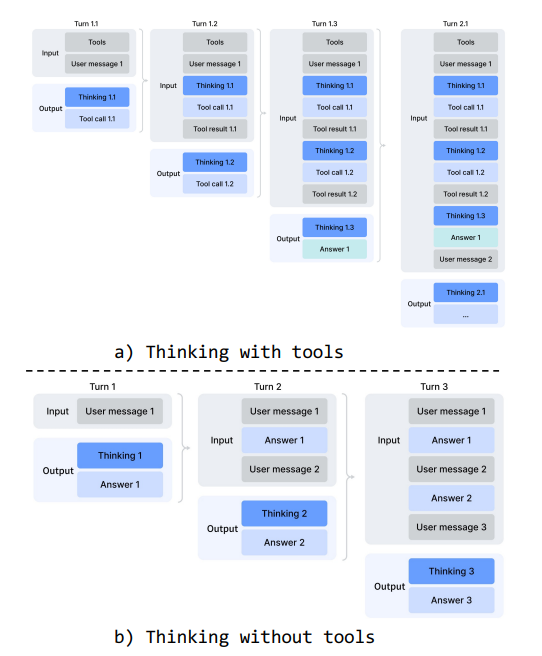}
  \caption{Figure 7 from the tech report of the DeepSeek-v4 \cite{DeepSeek} models.}
  \label{fig:deepseek}
\end{figure}

Figure \ref{fig:deepseek} shows figure 7 from the tech report for the DeepSeek-v4 \cite{DeepSeek} models. This visual rendition has striking resemblance to the way ACDL is rendered, but ACDL also formally captures the dynamic evolution using the time (turn) steps @T, the separation of messages, and the role of each message. 

The following is taken verbatim from the DeepSeek-V4 technical report \cite{DeepSeek}, which describes some of the context management strategies for their agent.
\begin{quote}
DeepSeek-V3.2 introduced a context management strategy that retains reasoning traces across tool-result rounds but discards them upon the arrival of new user messages. While effective, this still caused unnecessary token waste in complex agentic workflows
— each new user turn would flush all accumulated reasoning content, forcing the model to reconstruct its problem-solving state from scratch. Leveraging the expanded 1M-token context window of DeepSeek-V4 series, we further refine this mechanism to maximize the effectiveness of interleaved thinking in agentic environments:
\begin{itemize}
    \item \textbf{Tool-Calling Scenarios.} As illustrated in Figure 7(a), all reasoning content is fully preserved throughout the entire conversation. Unlike DeepSeek-V3.2, which discarded thinking traces upon each new user turn, DeepSeek-V4 series retain the complete reasoning history across all rounds, including across user message boundaries. This allows the model to maintain a coherent, cumulative chain of thought over long-horizon agent tasks.
    \item \textbf{General Conversational Scenarios.} As illustrated in Figure 7(b), the original strategy is preserved: reasoning content from previous turns is discarded when a new user message arrives, keeping the context concise for settings where persistent reasoning traces provide limited benefit.
\end{itemize}
\end{quote}

\begin{figure}[H]
  \centering
  \includegraphics[width=0.7\columnwidth]{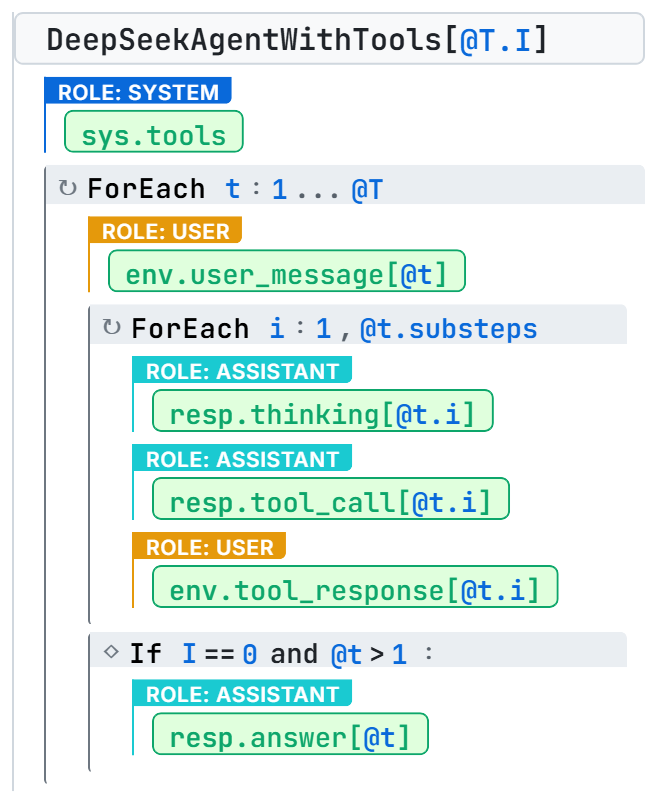}
  \caption{ACDL description of the DeepSeek agent with tools}
  \label{fig:deepseek-acdl-tools}
\end{figure}

\begin{figure}[H]
  \centering
  \includegraphics[width=0.7\columnwidth]{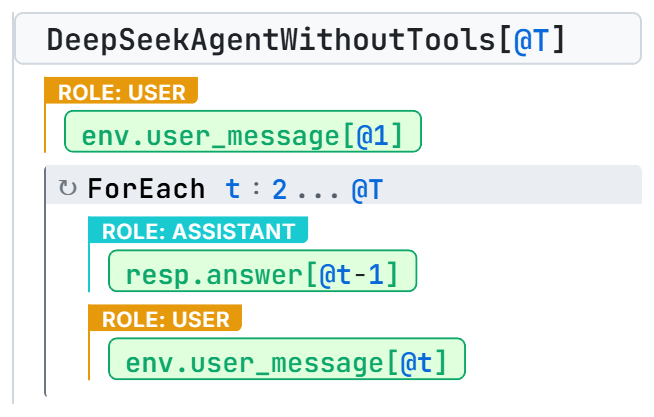}
  \caption{ACDL description of the DeepSeek agent without tools}
  \label{fig:deepseek-acdl-no-tools}
\end{figure}

Figures \ref{fig:deepseek-acdl-tools} and \ref{fig:deepseek-acdl-no-tools} show the ACDL specifications for the context descriptions shown in Figure \ref{fig:deepseek}. Our figures also state what the role is assigned to each of the messages. This might not be accurate for this specific case because this information is missing in the tech report description and is inferred by us.age 

\clearpage
\section{ACDL: Formal Language Reference}
\label{appendix:acdl}

This is a complete reference for the Agentic Context Description Language (ACDL) introduced in Section~\ref{sec:acdl}. The language declaratively specifies prompt structures sent to large language models, separating structural concerns from content and implementation.

\medskip\noindent\textbf{Specification Structure}

\noindent An ACDL specification defines a named prompt template parameterized by optional indices. The body consists of an ordered sequence of \emph{prompt blocks}---role messages, label blocks, and control flow constructs---which may be freely interleaved:

\begin{verbatim}
PromptName[idx1, idx2, ...]: {
  <prompt-blocks>
}
\end{verbatim}

\noindent A single file may contain multiple specifications, each defining a separate prompt template. Files may also contain \emph{fragment definitions}---reusable building blocks that encapsulate portions of a prompt. Fragments come in two varieties: string fragments (\texttt{StrFrag}), which produce content pieces, and role fragments (\texttt{RoleFrag}), which produce complete role messages. These are described in detail in the Fragments section below.

\begin{verbatim}
StrFrag FragmentName[params]: {
  <content-elements>
}

RoleFrag FragmentName[params]: {
  <prompt-blocks>
}

PromptName[idx1, idx2, ...]: {
  <prompt-blocks>
}
\end{verbatim}

\medskip\noindent\textbf{Role Messages}

\noindent Every message carries exactly one role. Four roles serve the chat format; a fifth pseudo-role serves the legacy completion format:

\smallskip\noindent
{\small
\begin{tabular}{@{}cp{7cm}@{}}
\toprule
\textbf{Marker} & \textbf{Purpose} \\
\midrule
\texttt{S:} & System---instructions, persona, tool descriptions, behavioral constraints. \\
\texttt{U:} & User---external input, observations, tool results. \\
\texttt{A:} & Assistant---prior model outputs, reasoning traces, chosen actions. \\
\texttt{T:} & Tool---structured tool call results. \\
\texttt{N:} & None---single unstructured text block (completion format only). \\
\bottomrule
\end{tabular}
}
\smallskip

\medskip\noindent Role messages support two syntactic forms. The \emph{multi-line} form encloses content in braces and permits any combination of content elements and control flow:

\begin{verbatim}
S: {
  INSTRUCTIONS
  AVAILABLE_TOOLS
  env.datetime[@t]
}
\end{verbatim}

\noindent The \emph{single-line} form omits braces and accepts exactly one content element---a context variable, template, or function call:

\begin{verbatim}
U: env.user_question[@t]
A: resp.answer[@t]
S: INSTRUCTIONS
\end{verbatim}

\noindent Control flow constructs (\texttt{ForEach}, \texttt{If}, \texttt{Switch}) are \emph{not} permitted in single-line role messages. To include control flow inside a role message, the multi-line braced form must be used:

\begin{verbatim}
U: {
  ForEach(item: env.items) {
    env.item_detail[@t, item]
  }
}
\end{verbatim}

\noindent The \texttt{N:} role (completion format) imposes two additional constraints: (1)~exactly one \texttt{N:} block may appear per prompt, and (2)~no chat roles (\texttt{S:}, \texttt{U:}, \texttt{A:}, \texttt{T:}) may appear in the same specification:

\begin{verbatim}
CompletionPrompt[@t]: {
  N: {
    TASK_DESCRIPTION
    env.context[@t]
    QUESTION
  }
}
\end{verbatim}

\medskip\noindent\textbf{Scoping Rules}

\noindent The language enforces a strict two-level scope that mirrors the structure of LLM chat APIs. At the \emph{top level} (the prompt body, outside any role block), only role messages, label and marker blocks, control flow constructs, name definitions, fragment invocations (role fragments), comments, and fragment definitions are permitted. \emph{Inside} a role block's braces, valid elements are context variables, functions, templates, control flow, comments, name definitions, fragment invocations (string fragments), marking blocks, and \texttt{break}/\texttt{continue}. Role messages may \textbf{not} appear inside other role messages---the following is invalid:

\begin{verbatim}
U: {
  S: INSTRUCTIONS   // ERROR: nested role
}
\end{verbatim}

\noindent For completion prompts using \texttt{N:}, the top level may contain only the single \texttt{N:} block---no other role messages, Mark blocks, or control flow may appear outside it.

\medskip\noindent\textbf{Context Variables}

\noindent Context variables reference dynamic runtime data. The general syntax is:

\smallskip
\noindent\texttt{namespace.path[indices]}

\smallskip\noindent where \texttt{namespace} is one of three reserved prefixes:

\smallskip\noindent
{\small
\begin{tabular}{@{}lp{7cm}@{}}
\toprule
\textbf{NameSpace} & \textbf{Use Cases} \\
\midrule
\texttt{env} & Environment---external inputs, observations, sensor readings, user queries, game state. \\
\midrule
\texttt{sys} & System---agent state, memory contents, tool configurations, action histories. \\
\midrule
\texttt{resp} & Response---prior LLM outputs, reasoning traces. \\
\bottomrule
\end{tabular}
}
\smallskip

\noindent Paths may be nested using dot notation to reach into sub-fields of structured data. Indices (described below) may appear at any level of the path:

{\footnotesize
\begin{verbatim}
env.user_question[@T]  // the user question at time t
sys.agent_desc // the agent description (constant, does not change over time)
sys.tool[@t].tool_response[@t]  // the tool response of tool at time t.
env.bomb_location[@T, bomb] // the bomb location of bomb at time T
\end{verbatim}
}

\noindent A variable without indices (e.g.\ \texttt{sys.agent\_desc}) refers to data that does not vary over time or other dimensions.

\medskip\noindent\textbf{Indices}

\noindent Indices address specific elements along one or more dimensions. Two types are distinguished syntactically.

The special symbol \texttt{@} denotes the primary time dimension that the prompt iterates over. What \texttt{@} represents depends on the agent's structure: for a ReAct agent that operates within a single turn but loops over many steps, \texttt{@} refers to the current step; for an agent that loops over multi-turn conversations, \texttt{@} refers to the current turn, and sub-steps within each turn are indexed with ordinary index variables.

The current time is denoted with capital letters: \texttt{@T} for the main time step, and \texttt{I}, \texttt{J}, etc.\ for sub-steps. When iterating over time steps, the corresponding lower-case letters are used. Sub-steps are accessed using dot notation: \texttt{@t.i} refers to sub-step \texttt{i} within turn \texttt{t}, while \texttt{@T.I} refers to the current sub-step of the current turn. When iterating over the sub-steps of a previous turn, use \texttt{@t.substeps} to obtain the count. For example:

\begin{verbatim}
@T            // Current time step
@T-1          // Previous time step
@T.I          // Current substep of current turn
@t.i          // Substep i of turn t (in loops)

// Iterating over all previous turns
ForEach(t: range(1, @T)) {
  env.observation[@t]
}

// Iterating over substeps in the current turn
ForEach(i: range(1, I)) {
  sys.action[@T.i]
}

// Nested: substeps within each previous turn
ForEach(@t: range(1, @T)) {
  ForEach(i: range(1, @t.substeps)) {
    sys.action[@t.i]
  }
}
\end{verbatim}

\smallskip
\noindent \emph{Non-time indices} have no prefix and address other dimensions---named entities, or context-variable-valued keys:

\begin{verbatim}
[sys.agent_name]
[bomb]

\end{verbatim}

\noindent Multiple indices are comma-separated: \texttt{env.bomb\_location[@t, bomb]} addresses a specific bomb at a specific time step. Standard arithmetic operators (\texttt{+}, \texttt{-}, \texttt{*}, \texttt{/}, \texttt{\%}) are permitted in all index positions, enabling expressions such as \texttt{@t-1}, \texttt{@t+1}, \texttt{t-k}, or \texttt{@t\,\%\,25}.

\medskip\noindent\textbf{Templates}

\noindent Templates are \texttt{ALL\_CAPS} placeholders representing text blocks whose content is specified at instantiation time. They describe the semantic purpose of a text section without fixing its wording, separating prompt architecture from prompt prose. Words within a template name are separated by underscores: \texttt{TASK\_INTRO}, \texttt{MAP\_DESCRIPTION}.

Templates may accept arguments in parentheses, enabling parameterized text:

\begin{verbatim}

INSTRUCTIONS        // Task explanation
AVAILABLE_TOOLS     // Tool list
QUERY(sys.agent_name)  // Depends on the agent's name

\end{verbatim}

\noindent An optional inline comment (after \texttt{//}) documents the intended content of the template.

\medskip\noindent\textbf{Functions}

\noindent Functions represent computed content---summarization, retrieval, formatting, or any transformation that cannot be expressed as a simple variable lookup. They are declared by name and purpose without defining their implementation; the name conveys semantic intent.

The syntax is \texttt{functionName(arg1, arg2, ...)[indices]}. The function names use \texttt{camelCase} (distinguishing them from \texttt{ALL\_CAPS} templates). Arguments may be context variables, time or regular indices, numeric literals, arithmetic expressions, or nested function calls:

\begin{verbatim}

summarize(sys.history[@t])
get_dialog_history(sys.agent_name)
range(1, @T, 2)

\end{verbatim}

\noindent The built-in \texttt{range(start, stop, step)} function generates numeric sequences for use in \texttt{ForEach} loops. The \texttt{step} argument is optional and defaults to~1. The range is exclusive, meaning \texttt{range(1, 3)} starts at~1 and ends at~2-1, excluding~3.

\medskip\noindent\textbf{Control Flow}

\noindent Three constructs govern dynamic prompt structure. All three may appear both at the top level (producing or gating entire role messages) and inside role blocks (controlling content within a single message).

\smallskip\noindent\emph{ForEach} iterates over ranges or collections to produce repeated structures. The syntax is:

\begin{verbatim}
ForEach(variable: iterable) {
  <body>
}
\end{verbatim}

\noindent The iterable may be a \texttt{range(start, stop, step)} call or a collection-valued context variable. At the top level, the loop body may contain role messages, producing multiple messages per iteration. Inside a role block, it produces repeated content elements:

\begin{verbatim}

// Top-level: produces role messages
ForEach(@t: range(1, @T)) {
  U: env.user_question[@t]
  A: resp.answer[@t]
}

// Inside a role: produces content
U: {
  ForEach(bomb: env.bombs) {
    env.bomb_location[@t, bomb]
    env.bomb_details[@t, bomb]
  }
}

// Collection iteration
ForEach(agent: sys.agent_names) {
  U: env.utterance[@t, agent]
}
\end{verbatim}

\smallskip\noindent\emph{If\,/\,ElseIf\,/\,Else} conditionally includes or excludes blocks based on runtime state. Conditions may use comparison operators (\texttt{==}, \texttt{!=}, \texttt{<}, \texttt{>}) and logical connectives (\texttt{\&}, \texttt{|}):

\begin{verbatim}
If @T > 1 {
  ForEach(t: range(1, @T)) {
    U: env.user_input[@t]
    A: resp.answer[@t]
  }
}

If sys.tool[@t] == get_clarification {
  U: env.user_input[@i]
}
ElseIf sys.tool[@t] == search {
  A: env.search_results[@t]
}
Else {
  A: sys.tool_used[@t].tool_response
}
\end{verbatim}

\smallskip\noindent\emph{Switch\,/\,Case\,/\,Default} selects among multiple alternatives based on the value of an expression:

\begin{verbatim}
Switch sys.action_type[@t] {
  Case "search" {
    U: env.search_results[@t]
  }
  Case "calculate" {
    U: env.calculation[@t]
  }
  Default {
    U: env.fallback[@t]
  }
}
\end{verbatim}

\noindent The \texttt{break} and \texttt{continue} keywords are available inside loops with their standard semantics.

\medskip\noindent\textbf{Early Termination}

The \texttt{PromptEndsHere when} construct signals that, if the given condition is true, the prompt ends at that point---no further content is appended to the message sequence sent to the LLM for that turn. This is useful when certain conditions require a truncated prompt, such as an initial turn that needs no history or context beyond the setup. The syntax is:

\begin{verbatim}
PromptEndsHere when <condition>
\end{verbatim}

\noindent For example, to end the prompt at the first sub-step of the current turn:

\begin{verbatim}
Prompt[@T]: {
  S: INSTRUCTIONS
  U: env.user_input[@T]
  PromptEndsHere when (@T == @t && @T.0)
  ForEach(i: range(1, @t.substeps)) {
    A: resp.answer[@T.i]
    U: env.feedback[@T.i]
  }
}
\end{verbatim}

\noindent Here, if the current time is the first substep of the current turn, the prompt contains only the system instructions and user input. Otherwise, the full history of substep interactions is appended.

\medskip\noindent\textbf{Fragments}

\noindent Fragments are reusable building blocks that encapsulate portions of a prompt specification. They enable modular prompt design by allowing common patterns to be defined once and invoked multiple times. Two kinds of fragments are supported, distinguished by what they produce when expanded.

\smallskip\noindent\emph{String Fragments} produce content pieces without an associated role. They are defined with the \texttt{StrFrag} keyword and may contain any elements valid inside a role block---context variables, functions, templates, control flow, other string fragments, name definitions, and comments. When invoked, the content expands in place and inherits the role of the enclosing message:

\begin{verbatim}
StrFrag DocumentContext[doc]: {
  env.doc_title[doc]
  env.doc_content[doc]
  summarize(env.doc_metadata[doc])
}

StrFrag ConversationContext[@T]: {
  CONTEXT_HEADER
  ForEach(@t: range(@T-5, @T)) {
    sys.Summary[@t]
    If sys.has_tool_call[@t] {
      sys.tool_response[@t]
    }
  }
}

\end{verbatim}

\noindent String fragments are invoked with the \texttt{Frag} keyword followed by the fragment name and arguments. They may appear anywhere a context variable, function, or template is valid---inside role blocks, within control flow bodies, or as arguments to other constructs:

\begin{verbatim}
U: {
  TASK_INSTRUCTIONS
  ForEach(doc: env.documents) {
    Frag DocumentContext[doc]
  }
  env.user_question[@T]
}
\end{verbatim}

\smallskip\noindent\emph{Roles Fragments} produce one or more complete role messages. They are defined with the \texttt{RolesFrag} keyword and may contain role messages, control flow, mark blocks, and other prompt-level constructs---the same elements valid at the top level of a prompt body:

\begin{verbatim}
RolesFrag ConversationTurn[@t]: {
  U: env.user_input[@t]
  A: resp.answer[@t]
  If sys.tool[@t] != none {
    T: sys.tool[@t].tool_response
  }
}
\end{verbatim}

\noindent Roles fragments are invoked at the top level of a prompt, wherever a role message would be valid. They expand to the full sequence of role messages defined in the fragment body:

\begin{verbatim}
ChatAgent[@T]: {
  S: INSTRUCTIONS
  ForEach(@t: range(1, @T)) {
    Frag ConversationTurn[@t]
  }
  U: env.user_input[@T]
}
\end{verbatim}

\noindent Both fragment types accept parameters, enabling parameterized reuse. Parameters follow the fragment name in square brackets and may include indices, context variables, or other valid index expressions. The same invocation syntax (\texttt{Frag Name[args]}) is used for both types; the parser determines which kind based on context---invocations inside role blocks resolve to string fragments, while those at the top level resolve to role fragments.

\smallskip\noindent Fragment definitions appear at the top level of an ACDL file, alongside prompt specifications. A single file may contain any combination of prompts and fragment definitions:

\begin{verbatim}
StrFrag ToolDescription[tool]: {
  sys.tool_name[tool]
  sys.tool_schema[tool]
}

RolesFrag ToolResult[@t, tool]: {
  A: sys.tool_call[@t, tool]
  T: sys.tool_response[@t, tool]
}

ToolAgent[@T]: {
  S: {
    INSTRUCTIONS
    ForEach(tool: sys.available_tools) {
      Frag ToolDescription[tool]
    }
  }
  ForEach(@t: range(1, @T)) {
    U: env.observation[@t]
    Frag ToolResult[@t, sys.selected_tool[@t]]
  }
  U: env.observation[@T]
}
\end{verbatim}

\vspace{1em}
\medskip\noindent\textbf{Mark Blocks}

A mark block annotates a section of the specification for visual emphasis in the rendered output. It places a bracket (\texttt{]}) along the side of the marked content, with a number identifier displayed beside it. Marks are purely presentational---they do not affect prompt semantics or scoping. They can wrap any prompt block, from a single content element to a large multi-message section:

\begin{verbatim}
Mark 1 {
  <prompt-blocks>
}
\end{verbatim}

\noindent The number appears next to the bracket in the visualization as \texttt{]1}. Multiple marks with different numbers may be used to highlight distinct sections:

\begin{verbatim}
Prompt[@T]: {
  Mark 1 {
    S: {
      INSTRUCTIONS
      AVAILABLE_TOOLS
    }
  }
  Mark 2 {
    ForEach(@t: range(1, @T)) {
      U: env.user_question[@t]
      A: resp.answer[@t]
    }
  }
  Mark 3 {
    U: env.user_question[@T]
  }
}
\end{verbatim}

\noindent Here, mark \texttt{]1} highlights the system setup, \texttt{]2} spans the conversation history loop, and \texttt{]3} marks the current query.

\medskip\noindent\textbf{Name Definitions}

A name definition binds a symbolic name to an expression, allowing complex or frequently repeated expressions to be written once and referenced concisely throughout the specification. The definition syntax is:

\begin{verbatim}
Name var_name := expression
\end{verbatim}

\noindent Once defined, the name is referenced using the \texttt{\$} prefix: \texttt{\$var\_name}. Name definitions improve readability by replacing long or opaque expressions with descriptive identifiers. The bound expression can be any valid ACDL element---a context variable, function call, arithmetic expression, or string literal. Fields of the bound value can be accessed via dot notation on the reference:

\begin{verbatim}
BasicRAG[@T]: {
  S: INSTRUCTIONS
  U: {
    Name docs := k_relevant_docs(env.user_input[@T])
    ForEach(i: range(1, $docs.len)) {
      $docs[i].source
      $docs[i].content
    }
    ANSWER_QUESTION_BASED_ON_DOCS
    env.user_input[@T]
  }
}
\end{verbatim}

\noindent Here, \texttt{\$docs} binds the result of a retrieval function, avoiding repetition of the full function call. Its length and individual elements are then accessed via \texttt{\$docs.len} and \texttt{\$docs[i]}.

Name definitions also support list comprehensions, which construct a list by iterating over a range or collection:

\begin{verbatim}
Name relevant_summaries :=
  [sys.summary[@t] for t in range(@T, @T-900, 100)]
compress_summaries($relevant_summaries)
\end{verbatim}

\noindent This binds \texttt{\$relevant\_summaries} to a list of summaries sampled every 100 steps, which is then passed as an argument to a function. The reference syntax is the same regardless of whether the bound value is a single expression or a list.

\medskip\noindent\textbf{Comments}

Comments use \texttt{//} syntax and may appear on any line---between prompt blocks, inside role blocks, or between specifications. Once a \texttt{//} is encountered, the remainder of that line is treated as a comment; no further ACDL elements may follow on the same line. An \emph{inline comment}, placed after a content element, renders alongside that element. A \emph{standalone comment}, placed on its own line, renders at the current level of nesting:

\begin{verbatim}
S: {
  // This comment appears on its own line,
  // indented to the level of the S: block
  INSTRUCTIONS  // Renders beside INSTRUCTIONS
  AVAILABLE_TOOLS
  // Another standalone comment
  env.datetime[@T]  // Renders beside datetime
}
// This comment is at the top level
ForEach(@t: range(1, @T)) {
  // This comment is inside the loop body
  U: env.user_question[@t]  // Beside the message
  A: resp.answer[@t]
}
\end{verbatim}

\medskip\noindent\textbf{Identifiers and Naming Conventions.}
Identifiers start with a letter or underscore and may contain letters, digits, and underscores. They are case-sensitive. The language enforces naming conventions to visually distinguish element types: templates use \texttt{ALL\_CAPS}, functions use \texttt{camelCase}, and context variable paths use \texttt{dot.separated.names}.

\medskip\noindent\textbf{An Illustrative Example}

Figure~\ref{fig:acdl-tool} presents a complete specification for a tool-using agent with conditional history replay. 

\begin{figure}[htbp]
\begin{verbatim}
ToolAgent[@T]: {
  S: {
    INSTRUCTIONS
    AVAILABLE_TOOLS
  }
  U: {
    env.user_input[@1]
    env.user_document[@1]
  }
  If t > 1 {
    ForEach(@t: range(1, @T)) {
      If sys.tool[@t] == get_clarification {
        U: env.user_input[@t]
      }
      Else {
        A: sys.tool[@t].tool_response
      }
    }
  }
  S: {REACT_INSTRUCTIONS}
}
\end{verbatim}
\caption{Tool-using agent. System messages frame the task; a conditional loop replays interaction history with role assignment determined by action type.}
\label{fig:acdl-tool}
\end{figure}

\end{document}